%% file: CVPR2024.tex
\documentclass[10pt,twocolumn,letterpaper]{article}

\usepackage[pagenumbers]{cvpr} % To force page numbers, e.g. for an arXiv version

\usepackage{algorithm}
\usepackage{algorithmic}
\usepackage{enumitem}
\usepackage{arydshln}
\usepackage{enumitem}
\newcommand{\openmatch}{\emph{OpenMatch}}
\newcommand{\openldn}{\emph{OpenLDN}}
\newcommand{\freematch}{\emph{FreeMatch}}
\newcommand{\flexmatch}{\emph{FlexMatch}}
\newcommand{\mixmatch}{\emph{MixMatch}}
\newcommand{\myparagraph}[1]{\smallskip\noindent\textbf{#1}}

\usepackage{caption}
\usepackage{subcaption}
\usepackage{color}
\usepackage{amsmath, dsfont}
\definecolor{airforceblue}{rgb}{0.36, 0.54, 0.66}
\definecolor{orange}{rgb}{1.0, 0.5, 0.0}

\newcommand{\app}{Suppl.}
\input{preamble}

\definecolor{cvprblue}{rgb}{0.21,0.49,0.74}
\usepackage[pagebackref,breaklinks,colorlinks,citecolor=cvprblue]{hyperref}

\def\paperID{6496} % *** Enter the Paper ID here
\def\confName{CVPR}
\def\confYear{2024}

\title{Semi-Supervised Learning in the Few-Shot Zero-Shot Scenario}%, With No Auxiliary Information}

\author {
    Noam Fluss,\textsuperscript{\rm 1}
    Guy Hacohen,\textsuperscript{\rm 1,2}
    Daphna Weinshall\textsuperscript{\rm 1} \\
    \textsuperscript{\rm 1} {School of Computer Science and Engineering, The Hebrew University of Jerusalem, Israel}\\
    \textsuperscript{\rm 2} {Edmond \& Lily Safra Center for Brain Sciences, The Hebrew University of Jerusalem, Israel}\\
    \{noam.fluss, guy.hacohen, daphna@mail.huji.ac.il\}
}

\begin{document}

\maketitle

\begin{abstract}
%Semi-Supervised Learning (SSL) leverages both labeled and unlabeled data to improve model performance. Traditional SSL methods assume that labeled and unlabeled data share the same label space. However, in real-world applications, especially when the labeled training set is small, there may be classes that are missing from the labeled set. To address this more general scenario, we propose a general approach to enhance existing SSL methods, such as \flexmatch, and allow them to handle this scenario by incorporating an additional term into their objective function. The additional term penalized for the KL divergence between the probability vectors of the true class probability vector and the inferred one. We demonstrate very large accuracy gains over state-of-the-art SSL, open-set SSL, and open-world SSL methods, on two benchmark image classification data sets, CIFAR-100 and STL-10. The gains are most pronounced when the labeled data is severely limited (1-25 labeled examples per class).
Semi-Supervised Learning (SSL) is a framework that utilizes both labeled and unlabeled data to enhance model performance. Conventional SSL methods operate under the assumption that labeled and unlabeled data share the same label space. However, in practical real-world scenarios, especially when the labeled training dataset is limited in size, some classes may be totally absent from the labeled set. To address this broader context, we propose a general approach to augment existing SSL methods, enabling them to effectively handle situations where certain classes are missing. This is achieved by introducing an additional term into their objective function, which penalizes the KL-divergence between the probability vectors of the true class frequencies and the inferred class frequencies. Our experimental results reveal significant improvements in accuracy when compared to state-of-the-art SSL, open-set SSL, and open-world SSL methods. We conducted these experiments on two benchmark image classification datasets, CIFAR-100 and STL-10, with the most remarkable improvements observed when the labeled data is severely limited, with only a few labeled examples per class
\end{abstract}

\section{Introduction}

\begin{figure}[t!]
    \begin{subfigure}{.49\columnwidth}
      \centering
      \includegraphics[width=.9\linewidth]{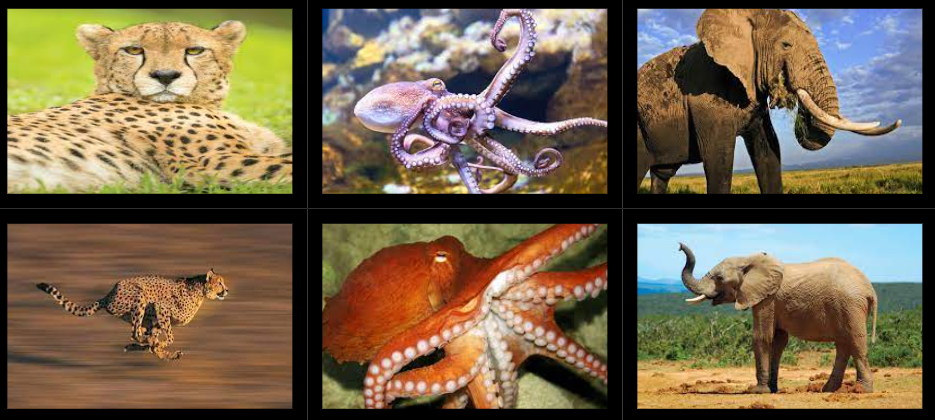}
%     \vspace{-0.62cm}
      \caption{Labeled train data}
     \vspace{0.2cm}
%      \label{subfig:approaches_1}
    \end{subfigure}
    \begin{subfigure}{.49\columnwidth}
      \centering
      \includegraphics[width=.9\linewidth]{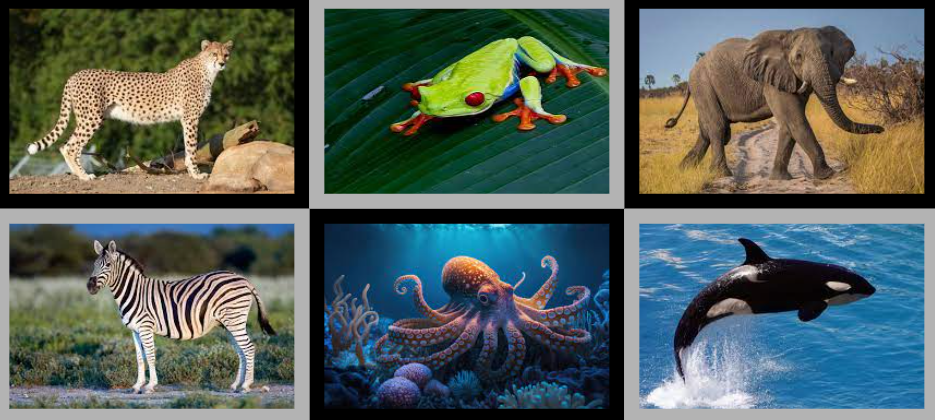}
%     \vspace{-0.62cm}
      \caption{Unlabeled train data}
      \vspace{0.2cm}
%      \label{subfig:approaches_2}
    \end{subfigure}
    \begin{subfigure}{\columnwidth}
      \centering
      \includegraphics[width=.85\linewidth]{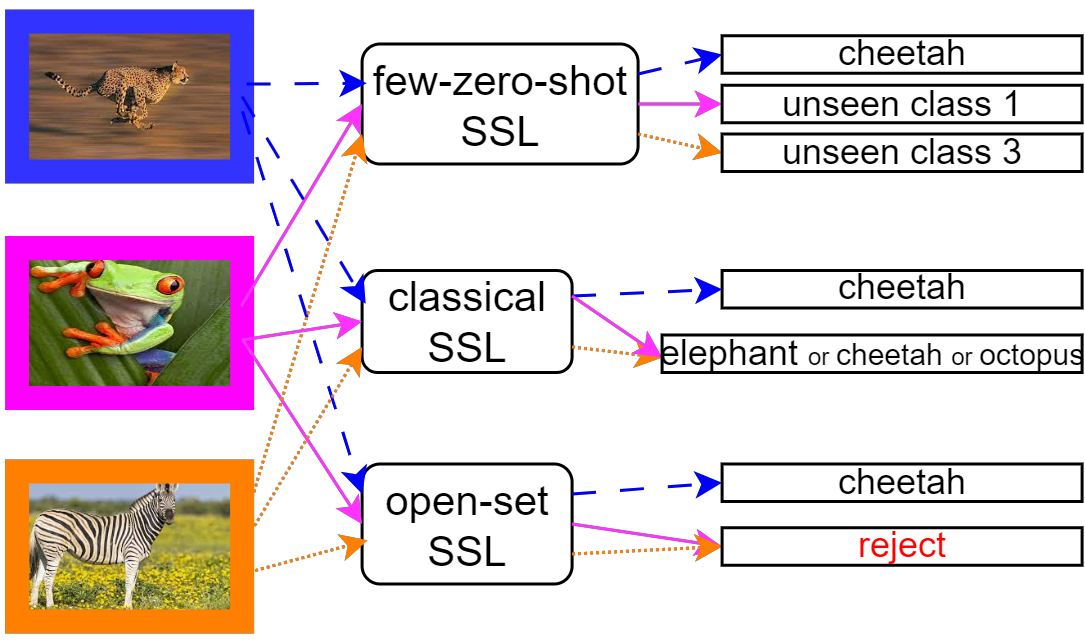}
%    \caption{}
%      \label{subfig:approaches_3}
    \end{subfigure}
      \vspace{-0.42cm}
    \caption{A toy example with 6 classes: 3 seen  \{cheeta, elephant, octopus\}, and 3 unseen  \{zebra, whale,  frog\}. (a)-(b) Images from seen classes are marked by black bounding boxes, while images from unseen classes have grey bounding boxes. Bottom: Different classification outcomes when adopting different SSL frameworks.}
    \label{fig:frameworks}
      \vspace{-0.4cm}
\end{figure}

As the collection of data becomes increasingly widespread, new scenarios that pose challenges to the traditional supervised machine learning framework are continuously emerging. Consider, for instance, a scenario %where data can be easily collected at a low cost. However, acquiring 
wherein the acquisition of accurate labels for data collected in an unsupervised manner requires specialized expertise or expensive machinery. Such scenarios may occur more frequently in fields like medical research or genomics. In this scenario, the size of the unlabeled dataset can be vast, but due to limited resources, only a small portion of it may be selected in an unsupervised manner to be further processed for labeling\footnote{We note that active learning aims to replace the random selection of the set of points to be labeled by a more effective selection methodology.}. In the end, we are left with a substantial unlabeled dataset and a relatively limited number of labeled data points. 

The above scenario is unique in that by assumption, the composition of the set of labeled examples is not monitored by a teacher (or oracle). In particular, when the labeled set is very small, certain classes may be totally absent from the labeled set. We call them ``zero-shot" or ``unseen" classes. %Note also that the small labeled set is unlikely to faithfully represent the prior probability distribution of the classes. 
The remaining classes are called ``few-shot" or ``seen". Thus, in the proposed scenario, the labeled set is constructed of classes with only a few labeled examples, or none at all. Additionally, due to the random selection of points to be processed for labeling and differently from traditional few-shot scenarios, the class distribution of labeled points may deviate widely from the true class distribution. 

\begin{table*}[ht]
\small
\centering
\resizebox{\textwidth}{!}{
%    \begin{tabular}{>{\raggedright}m:{3cm} c: c  c: c  c c  c c}
    \begin{tabular}{l : c c : c c : c c : c c}
    \toprule
     & \multicolumn{2}{c:}{\textbf{Train labeled data}} & \multicolumn{2}{c:}{\textbf{Train unlabeled data}} & \multicolumn{2}{c:}{\textbf{Test data}} & \multicolumn{2}{c}{\textbf{Expected output}} \\
    \cmidrule(lr){2-3} \cmidrule(lr){4-5} \cmidrule(lr){6-7} \cmidrule(lr){8-9}
    \textbf{Settings} & \textbf{seen} & \textbf{unseen} & \textbf{seen} & \textbf{unseen} & \textbf{seen} & \textbf{unseen} & \textbf{seen} & \textbf{unseen} \\
    \midrule
    $\dagger$zero-shot learning & Y & N  & - & -  & Y/N & Y  & classify & classify \\ \midrule
    *Semi-Supervised Learning (SSL) & Y & -  & Y & Y/N &  Y & -  & classify & -  \\ \hdashline
    *Open set SSL (OSSL) & Y & N & Y & Y  & Y & Y  & classify & reject \\ \hdashline
    \hspace{3.2pt} Open world SSL (OWSSL)  & Y & N  & Y  & Y &  Y & Y  & classify& discover/classify\\ 
    \midrule
    \hspace{6.5pt}\textbf{Our settings} & few-shot & N & Y & Y & Y & Y & classify & classify \\
    \bottomrule
    \end{tabular}
}
    \caption{Catalogue of different relevant frameworks as discussed in the text. The training set is divided into seen and unseen classes, a distinction not originally made in most frameworks. For each framework, the symbols 'Y' and 'N' denote that such data as specified in the column is permitted in the framework; 'Y' indicates that such data exists, and 'N' that it doesn't. '*' indicates frameworks wherein unseen classes may include OOD samples (see text). %'Scarce' indicates that seen classes may have only a few examples. 
    $\dagger$ indicates that additional semantic side-information such as attributes is needed. %When the algorithm makes use of the test set (transductive setting), 'same as test' is indicated.
    \textbf{Expected output:} 
    Open-set SSL methods are not designed to classify new examples from unseen classes, only to reject them. Most open-world SSL methods are transductive, designed to 'discover' structure in a fixed test set rather than classify new unseen examples. }
    \label{tab:updated_table}
\end{table*}

We identify valuable constraints in this scenario, called \emph{few-zero-shot SSL} in Fig.~\ref{fig:frameworks}, which can be leveraged for the successful handling of zero-shot classes. Firstly, by assumption the unlabeled training set is derived from the same distribution as the test dataset. % , which consists of new samples obtained under similar conditions.
Secondly, there is no need to account for Out-Of-Distribution (OOD) samples. Finally, the set of possible labels is fixed throughout.

While Semi-Supervised Learning (SSL) appears to be a suitable framework to address the problem at hand, it is important to note that most SSL methods assume the presence of all classes in the labeled set, namely, that there are no zero-shot classes. A step in the right direction is taken in the \emph{open-set SSL} (OSSL) framework, where the data contains points sampled from either unseen classes or OOD (or both). However, open-set techniques are designed to reject test samples that do not belong to any seen class and do not predict missing class labels. 

These differences between frameworks are illustrated in Fig.~\ref{fig:frameworks} and Table~\ref{tab:updated_table}. To evaluate the significance of these differences, we compare our proposed method to state-of-the-art SSL and OSSL methods adapted to identify unseen classes (see details in Section~\ref{sec:adapt}), showing large performance gains. % for our approach. 

A closely related scenario is addressed in the \emph{open-world SSL} framework, which aims to classify seen classes \textbf{and} to discover unseen classes in an unlabeled set. Typically in this framework, a learner is given substantial labeled and unlabeled sets for training, where both sets are sampled from the same distribution. Originally, the aim of such methods was to partition the unlabeled set in a transductive manner. Such methods can be extended to our inductive scenario by providing the partitioned unlabeled set as input to a regular SSL method, see discussion of related work below.
% may be irrlevent - \noam{the unlabeled sets is not bigger than the one we use - it's standard. may be we should change the sentence to: "a learner is provided with an unlabeled set alongside a sizable labeled set for training" \daphna{where do we say that it is bigger? why do you make this comment? in the future, when you add comments, make sure you don't merge paragraphs, as you have done here until I fixed it}}

Another related methodology is Zero-Shot Learning (ZSL). Traditionally, ZSL methods are designed to train models that can classify objects from unseen classes by utilizing semantic side-information and knowledge acquired from seen classes \citep[see survey by][]{pourpanah2022review}. Differently, in our scenario, we lack access to any auxiliary semantic information. Another difficulty with adopting this framework to our scenario stems from the fact that ZSL usually assumes that seen classes are adequately sampled, i.e., they are \textbf{not} ``few-shot" but ``many-shot". 

\subsection{Our proposed approach: outline} 

We propose a simple end-to-end framework that tackles the few-zero-shot SSL scenario, offering a general solution that can be used to enhance any semi-supervised learning (SSL) method. The approach consists of two main steps: (i) Modify the last layer of a given SSL architecture to incorporate both seen and unseen classes. By doing so, the model is capable of handling and classifying instances from both types of classes. (ii) Introduce an additional term into the loss function, which penalizes discrepancies between the output class-distribution of the trained model and the veridical class-distribution. 

As the field of semi-supervised learning continues to evolve, the proposed approach can be used to adapt future SSL methods to our scenario. When compared to zero-shot learning, instead of relying on hard-to-collect semantic side information, we exploit the fact that the unlabeled set accurately represents the underlying class distribution and contains classes that are not present in the labeled set. By doing so, we are able to transfer knowledge from few-shot to zero-shot classes without auxiliary information. 

We assess the performance of our method on standard image classification datasets, re-constructing training sets where labels are obtained as described above. While varying the number of unseen classes, we also vary the conditions under which seen classes are sampled, including both balanced and imbalanced scenarios. Current SSL methods, after being adapted to identify unseen classes (see details in Section~\ref{sec:adapt}), are unable to successfully classify them. Our methodology, when incorporated with the same SSL methods, greatly improves their performance on unseen classes, while usually matching or even surpassing their performance on seen classes. Additionally, our method surpasses the overall performance of both open-set and open-world SSL methods, see Section~\ref{sec:results}. The improvement is especially pronounced when the labeled set is small.

\subsection{Related work}

A number of related frameworks have been investigated in the machine learning community, motivated by problems that also inflict our scenario. Table~\ref{tab:updated_table} summarizes the \emph{main characteristics of the most relevant frameworks}, highlighting the differences between them. Next, we discuss each one in more detail.

\paragraph{Closed- and Open-Set Semi-Supervised Learning,} 
denoted \emph{SSL} and \emph{OSSL} respectively. SSL has made remarkable progress in recent years, see survey by \citet{chen2022semi}. It involves leveraging sparsely labeled data and a considerable amount of additional unlabeled data, typically sampled from the same underlying data distribution as the labeled data. In more general settings, Out Of Distribution (OOD) data may also be present in the unlabeled set.

In the original SSL settings \citep{chapelle2009semi}, both the labeled and unlabeled data share a label set within the same domain. It is not, however, suitable for our scenario because it does not allow for unseen classes. Going a step further, open-set SSL (OSSL) permits the presence of unseen classes or OOD data in the unlabeled set. However, this framework is only designed to identify and reject data points that do not belong to the classes in the labeled set, and is not tailored to classify unseen classes.

Some of the most effective SSL methods at present time combine ideas from self-supervised learning (e.g., consistency regularization), data augmentation, and the assignment of pseudo-labels to unlabeled data points, in order to utilize unlabeled data \citep[e.g.,][]{DBLP:conf/aaai/ChenZLG20,sohn2020fixmatch,DBLP:conf/nips/ZhangWHWWOS21,DBLP:conf/iclr/Wang0HHFW0SSRS023}. The field of Open-set SSL (OSSL) is also rapidly developing \citep{guo2020safe,yu2020multi,huang2021trash,park2022opencos,DBLP:journals/tmlr/SunL23}. For comparison, we use the competitive OSSL method proposed by \citet{DBLP:journals/corr/abs-2105-14148}, which uses the SSL and novelty detection frameworks in combination. 

\paragraph{Open-World SSL,} denoted \emph{OWSSL}. This paradigm is related to \emph{Novel Class Discovery}, which aims to discover unseen classes within unlabeled data \citep{xie2016unsupervised,han2020automatically,wang2020open}. Joining SSL with this paradigm, OWSSL \citep{cao2022open,zhong2021openmix,rizve2022openldn,guo2022robust} aims to discover unseen classes in an unlabeled set and to classify seen classes. Initially, OWSSL methods were transductive, in the sense that the test set is processed as a single batch, and classes are discovered by partitioning this set. Typically this is accomplished by learning pairwise similarities between the points in the unlabeled set.

In \openldn, \citet{rizve2022openldn} extended the scope of the open-world methodology to our scenario. Specifically, they proposed to add a step to the algorithm and use its outcome, both seen classes and the partitioned unlabeled set, as input to train a new SSL model, thus obtaining a classifier for both seen and unseen classes (the latter up to permutation of unseen labels). Our method differs from \openldn\  in that it solves the problem in an end-to-end fashion, and is thus more suitable for the low-budget regime with very few labeled examples. 

\paragraph{Zero-Shot Learning,}
denoted \emph{ZSL}. Work on ZSL often involves the computation of a separate label embedding space and assumes access to semantic data that may be difficult to obtain. Some recent work aims to expand the scope of this methodology to domains where the seen classes are only sparsely sampled. For example, \citet{li2015semi,xu2021semi} leverage SSL in order to aid the computation of label embedding, in a two-step manner. 

\subsection{Summary of contribution}
We introduce and motivate a novel SSL scenario, characterized by classes that are present in both the unlabeled set and the test set, yet absent from the labeled set. Addressing this scenario, we offer a solution that:

\begin{enumerate}[label=\roman*]
    \item Outperforms all current baselines by a large margin, including SOTA methods from related SSL frameworks.
    \item Can be assimilated into any SOTA SSL method.
    \item Converges significantly faster than its underlying SSL method in few-shot settings.
\end{enumerate}

% \begin{enumerate}
%     \item Outperforms all current baselines by a large margin.
%     \item Can be assimilated into any state-of-the-art SSL method.
%     \item In few-shot settings, the combined solution converges significantly faster than its underlying SSL method.
% \end{enumerate}

\section{Our Approach}

%Traditionally, different metrics are used to evaluate performance when dealing with labeled or unlabeled data. With labeled data, metrics that measure classification accuracy are appropriate. With unlabeled data, the outcome is essentially a partition of the data -- no specific labels are assigned to specific clusters. Here, the appropriate metrics measure the partition accuracy, as used to evaluate clustering algorithms. These scores are borrowed from the domains of novel class discovery and open-world SSL, and are detailed in Section~\ref{sec:scores}.

After the introduction of some notations, in Section~\ref{sec:ourmethod} we describe our method, designed to learn a classifier for the Semi-Supervised Learning scenario with few-shot and zero-shot classes. In Section~\ref{sec:adapt} we discuss how baseline methods can be adapted and used for comparison. Specifically, since most of the relevant frameworks (see discussion above) are not designed to classify unseen classes, we describe an adaptation step, which allows us to obtain unseen class classification from SSL and open-set SSL methods. Finally, in Section~\ref{sec:scores} we detail the different metrics that are used to evaluate performance when dealing with labeled or unlabeled data, as customary.

\subsection{Notations and definitions}

Let $\mathcal{X}$ denote the set of all possible examples, and $C$ the set of possible classes. We consider a learner denoted $f:\mathcal{X}\rightarrow C$, which has access to a labeled set of examples $\mathcal{L}\subseteq\mathcal{X}\times C$ and an unlabeled set of examples $\mathcal{U}\subseteq\mathcal{X}$. In our scenario, the labeled examples in $\mathcal{L}$ do not contain all possible classes of $C$. Let $C_{seen}\subseteq C$ denote the set of classes that appears in $\mathcal{L}$, and $C_{unseen}\subseteq C$ the set of classes that do not appear in $\mathcal{L}$. Note that $C_{seen}\cap C_{unseen} =\emptyset$, and $C_{seen}\cup C_{unseen} =C$. 

For evaluation purposes, we assume the existence of a test set $\mathcal{T}\subseteq\mathcal{X}\times C$, which is disjoint from both $\mathcal{U}$ and $\mathcal{L}$. $\mathcal{T}$ contains examples from all classes. We denote the examples from seen classes in the test set as $\mathcal{T}_{seen}$ and the examples from unseen classes as $\mathcal{T}_{unseen}$.   

\smallskip\noindent
\textbf{Empirical class probability, definition.}
Let $B$ denote the batch size of the learner and $[\overline{p}_i]_{i\in C}$ the vector of empirical class probability. In a specific batch $\{x_b\}_{b=1}^B$, $\overline{p}_i$ is computed as follows:
\begin{equation*}
\overline{p}_i = \frac{1}{B}\sum_{b=1}^{B} f_i(x_b),  \quad {i\in C}
\end{equation*}
Above $f_i(x_b)$ denotes the soft-max confidence of learner $f$ in the assignment of example $x_b$ to class $i\in C$. Finally, let $[q_i]_{i\in C}$ denote the vector of veridical marginal class probability derived from the data distribution over $\mathcal{X}\times C$:
\begin{equation*}
q_i = E_{x\in\mathcal{X}}[\mathds{1}_{[label(x)=i]}],~~i\in C
\end{equation*}

\subsection{Our Method}
\label{sec:ourmethod}

\subsubsection*{Objective function}

%\noam{https://openreview.net/pdf?id=6yVvwR9H9Oj}\\
Let $\ell_{kl}$ denote the KL-divergence between the two probability vectors $[q_i]_{i\in C}$ and $[\overline{p}_i]_{i\in C}$:
\begin{equation}
\label{eq:l_kl}
\ell_{kl} = \sum_{i\in C_{seen}\cup C_{unseen}} \overline{p}_i \left [ \log(q_i)  -  \log(\overline{p}_i) \right ]
\end{equation}
If the prior vector of class frequencies $[q_i]_{i\in C}$ is uniform (i.e., all classes are equally likely), this definition reduces to the negative entropy of the empirical class frequencies (up to an additive constant):
\begin{equation*}
\ell_{kl} = \frac{1}{|C|}-\sum_{i\in C_{seen}\cup C_{unseen}} \overline{p}_i \log(\overline{p}_i)
\end{equation*}

\begin{figure}[htb!]
      \centering
      \includegraphics[width=.85\linewidth]{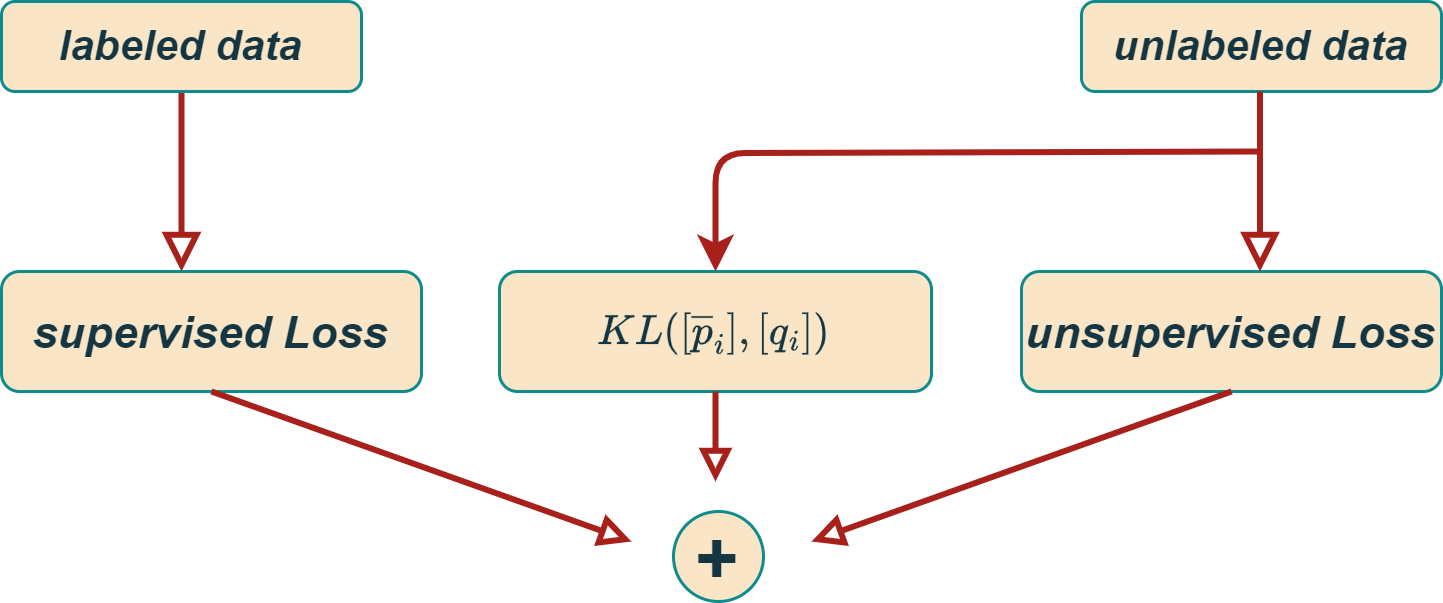}
    \caption{The combined SSL loss used in our approach, where $KL([\overline{p}_i],[q_i])$ denotes the KL-divergence between the vectors.}
    \vspace{-0.5cm}
    \label{fig:method}
\end{figure}

Most SSL methods optimize a loss function which is essentially the sum of an unsupervised term $\ell_{u}$ and supervised term $\ell_{s}$. In our approach, we adopt the full pipeline of the method, while optimizing the following objective function:
\begin{equation*}
\ell = \ell_{s} + \ell_{u} +  \lambda \ell_{kl}
\end{equation*}
This design is illustrated in Fig.~\ref{fig:method}. 

\subsubsection*{Early stopping}

During the training of our algorithm, we sometimes encounter a phenomenon in which the confidence of the model's predictions falls drastically, especially the confidence of points from unseen classes.
This phenomenon results in the model being essentially unable to generate pseudo labels, and therefore the scores show a sharp sudden decline. This may happen relatively early in the training as there are fewer labels per class. To address this problem, we track the gradient graph of the loss term $\ell_{kl}$ during training. Once a sharp decline is identified, the training is stopped a few epochs prior to the sudden decline. Interestingly, this procedure has a curious artificat, where our algorithm converges much faster than its original SSL variant while achieving a greatly improved accuracy. %For example, with 4 labels per class and 40 unseen classes in CIFAR-100, our method converges after $523.0 \pm 24.61$ epochs, while the SSL methods of \flexmatch\  and \freematch\  require 1024 epochs, \openmatch\  - 512, and \openldn\  - 1024.

\subsection{Baseline methods for comparison}
\label{sec:adapt}

Semi-Supervised Learning in the Few-Shot Zero-Shot Scenario is a new setting, and there are no ready-to-use baseline methods that are designed to solve it effectively. Consequently, we extend methods designed to deal with closely related problems, as discussed in the introduction (see Table~\ref{tab:updated_table}), and report their results. It should be kept in mind that since the methods are designed for other scenarios, their inferior results here should not be taken as evidence that they are not suited for their original task, where they achieve state-of-the-art results. 

More specifically, we adapt methods designed to deal with the following scenarios: (i) traditional semi-supervised learning (SSL), (ii) open-set SSL (OSSL), (iii) open-world SSL (OWSSL). We start by describing how to extend SSL and OSSL methods to the OWSSL scenario, similarly to how it's done by \citet{cao2022open}. We then conclude by discussing the adaptation of the OWSSL scenario.

\smallskip\noindent
\textbf{Traditional SSL} \hspace{0.1cm}
These methods are designed to classify seen classes, but they do not expect to encounter unseen classes in the test set. In order to extend them to the OSSL scenario, we adopt a common way to add a reject option to classifiers \citep[e.g.,][]{tax2008growing}, by applying a threshold test to the softmax confidence score of the model. To compensate for the simplicity of the approach, we choose for each method its optimal threshold based on ground truth information. In the end, we have a set of points classified into seen classes, and a set of rejected points, similar to the OSSL scenario to be discussed next.

\smallskip\noindent
\textbf{Open-set SSL} \hspace{0.07cm}
These methods are designed to classify seen classes and reject unseen classes, but do not classify unseen classes. In order to extend these methods to our scenario, an additional step is employed to partition the rejected points to $|C_{unseen}|$ parts. This is accomplished with K-means clustering, performed over the model's feature space. 

As customary for unlabeled data, and for the purpose of evaluation only, the set of cluster labels is matched to the set of unseen labels $C_{unseen}$ using the best permutation. Note that the \textit{unseen classes accuracy} score in (\ref{eq:acc_unseen}) below involves a similar procedure. As a result, the set of test points is now classified to all the labels in $C$.

\smallskip\noindent
\textbf{Open-world SSL} \hspace{0.1cm}
These methods are designed to process the labeled and unlabeled training sets and output a partition of the unlabeled set. This partition is matched to $C_{unseen}$ as described above. Finally, in order to generate a classifier for future unseen points, we follow the procedure suggested by \citet{rizve2022openldn}: use the labeled training set as is and the unlabeled set with its inferred labels to train another SSL classifier, whose domain of output labels is $C$. Note that the outcome of the two-stage model is unique up to permutation of the unseen labels $C_{unseen}$.

\subsection{Evaluation scores}
\label{sec:scores}
    
%\paragraph{Open-world SSL scores.}
As customary in novel class discovery and OWSSL, the following scores are used:
\begin{itemize}
    \item \textit{Seen classes accuracy}
    \begin{equation}
    \label{eq:acc_seen}
      acc_{seen}=\frac{1}{|\mathcal{T}_{seen}|}\sum_{x_i\in \mathcal{T}_{seen}}\mathds{1}_{[f(x_i)=y_i]}  
    \end{equation}
    \item \textit{Unseen classes accuracy}
    \begin{equation}
    \label{eq:acc_unseen}
      acc_{unseen}=\frac{1}{|\mathcal{T}_{\scriptscriptstyle unseen}|} \sum_{x_i\in \mathcal{T}_{unseen}}\mathds{1}_{[\tilde f(x_i)=y_i]}  
    \end{equation}
    Above $\tilde f(x_i)$ denotes the class assignment of $x_i$ based on the best permutation of the unseen classes, as customary when measuring clustering accuracy.
    \item
    \textit{Combined score}\\ %Accumulated score, accuracy of seen + clustering of unseen
    \begin{equation}
    \label{eq:acc_combined}
    \frac{|C_{seen}|}{C} \cdot acc_{seen}+ \frac{|C_{unseen}|}{C}\cdot acc_{unseen}
    \end{equation}
    \end{itemize}

\section{Empirical Evaluation}
\label{sec:empirical-evaluation}

\begin{figure*}[htb!]
\centering
    \begin{subfigure}{.33\textwidth}
    \centering
      \includegraphics[width=1\linewidth]{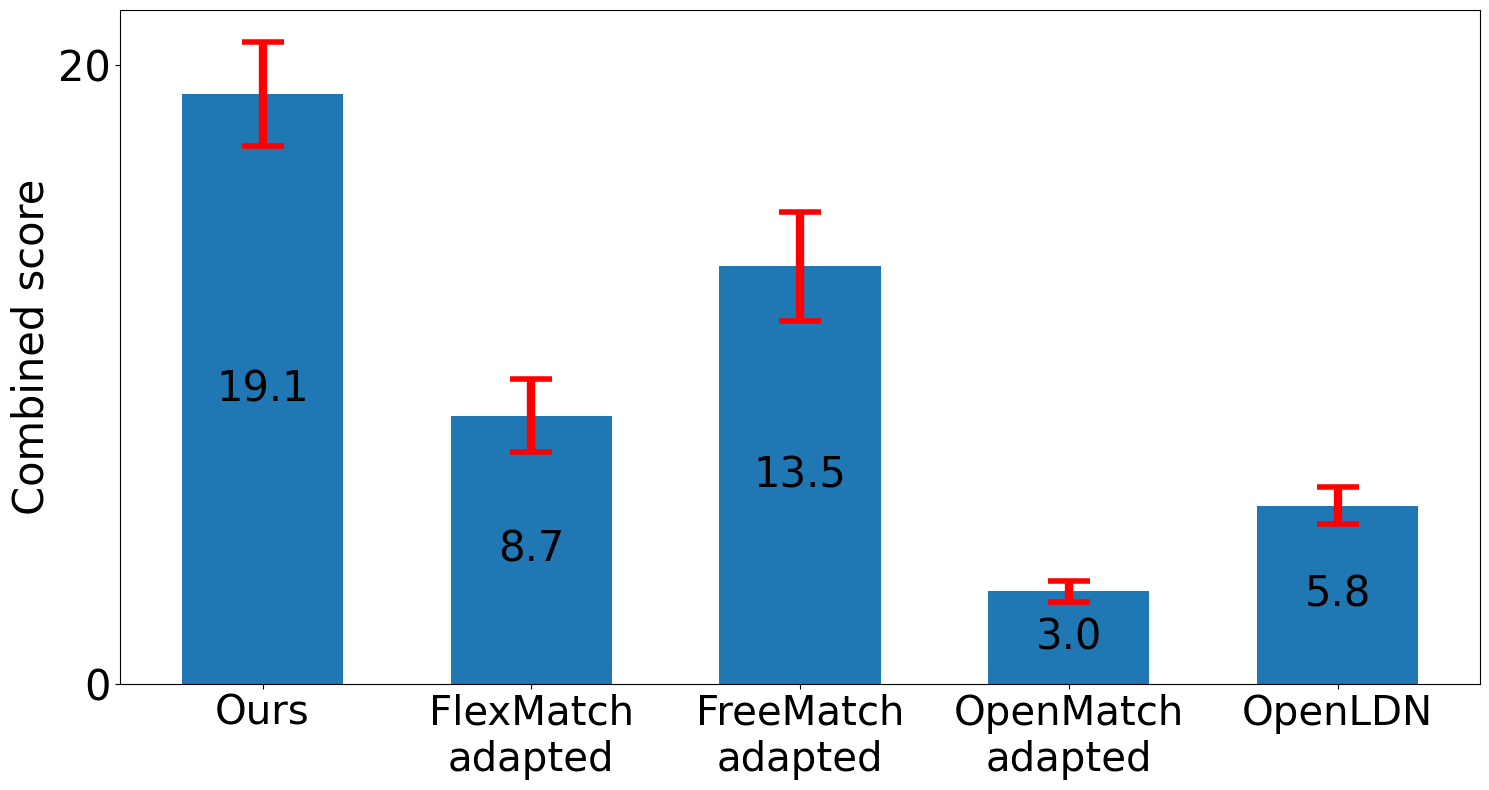}
      % \vspace{-0.62cm}
      % \caption{\textit{Combined score}}
      % \vspace{0.2cm}
      % \label{subfig:1_labels_per_class_35_unseen_wrn_28_2_combine}
    \end{subfigure}
    \begin{subfigure}{.33\textwidth}
    \centering
    % \textbf{4 labels per class, 40 unseen classes}
      \includegraphics[width=1\linewidth]{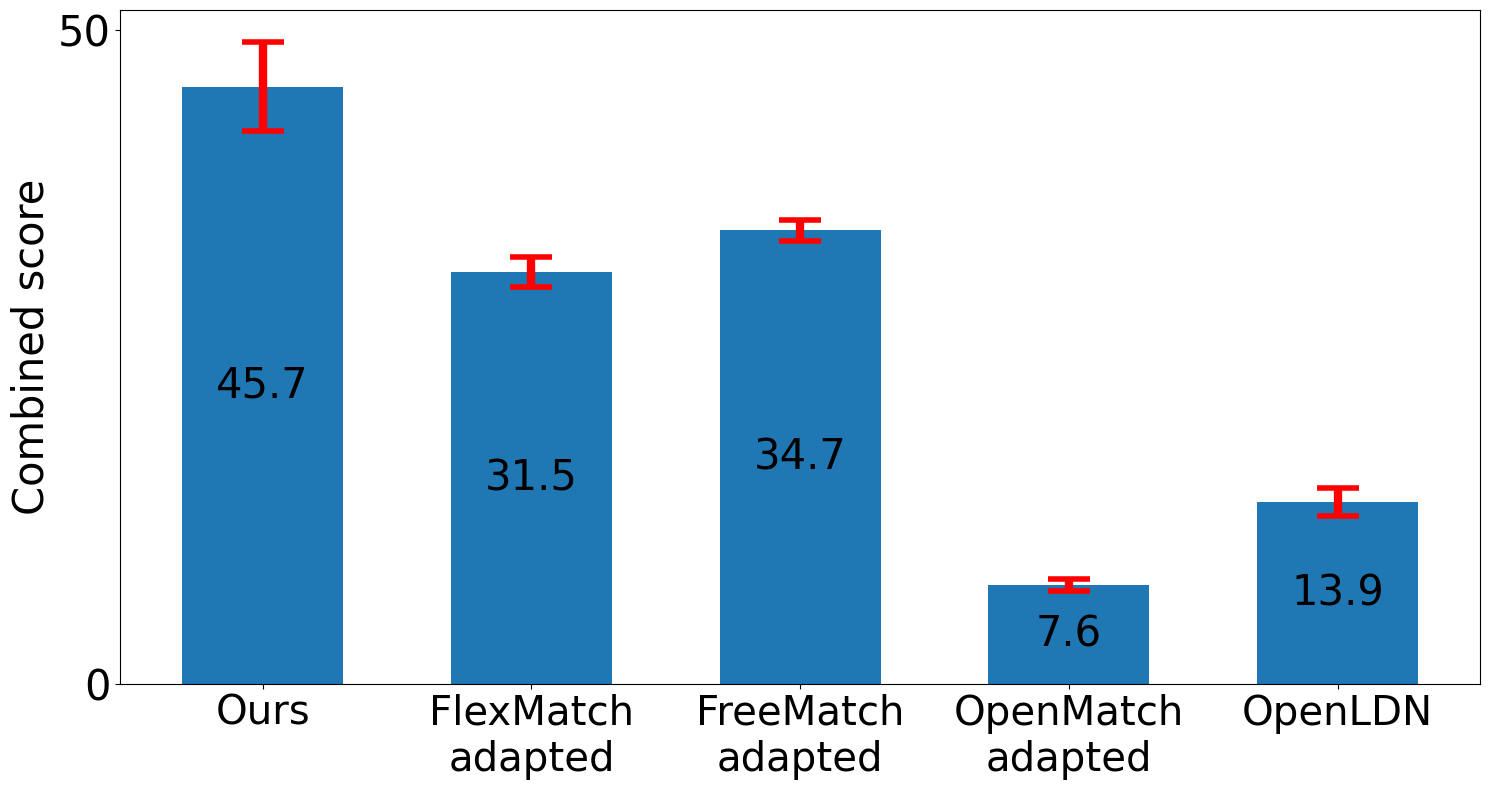}
      % \caption{\textit{Combined score}}
      % \label{subfig:4_labels_per_class_40_unseen_wrn_28_8_combine}
    \end{subfigure}
    \begin{subfigure}{.33\textwidth}
    \centering
    % \textbf{25 labels per class, 40 unseen classes}
      \includegraphics[width=1\linewidth]{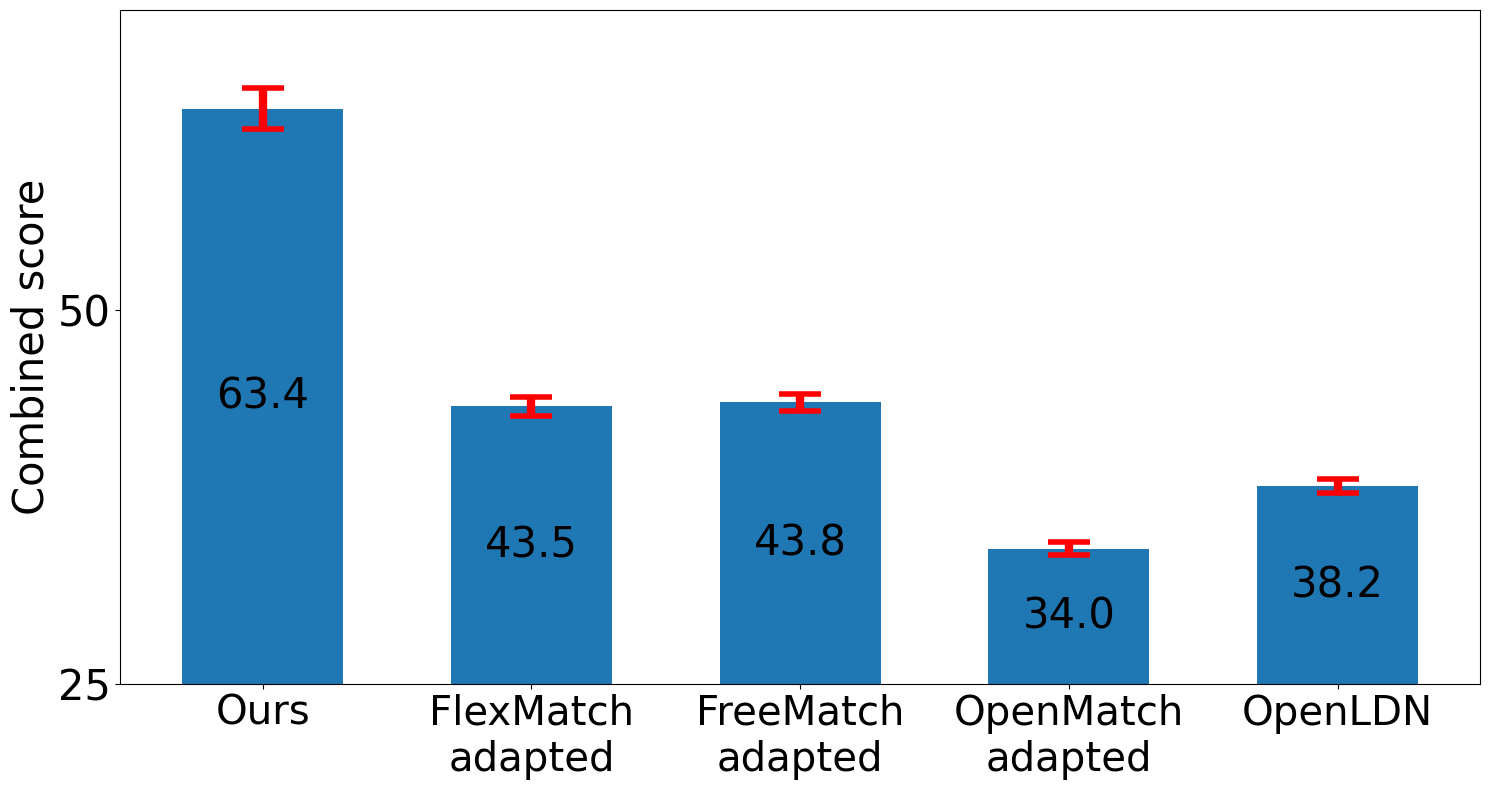}
      % \vspace{-0.62cm}
      % \caption{\textit{Combined score}}
      % \vspace{0.2cm}
      % \label{subfig:25_labels_per_class_40_unseen_wrn_28_8_combine}
    \end{subfigure}
    
    \begin{subfigure}{.33\textwidth}
      \centering
      \includegraphics[width=1\linewidth]{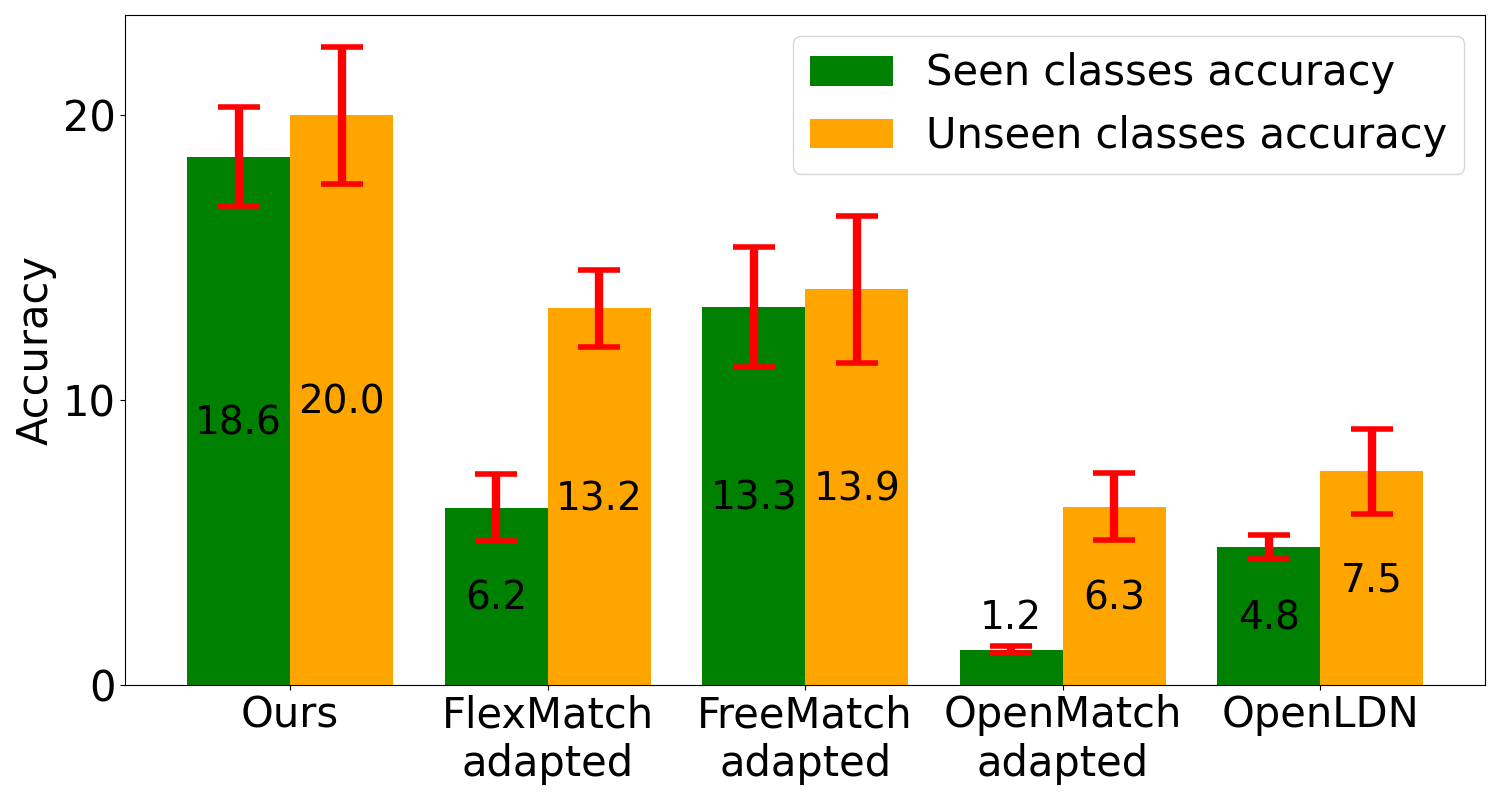}
      % \vspace{-0.62cm}
      \caption{1 label per class, 35 unseen classes}
      \label{subfig:1_labels_per_class_35_unseen}
    \end{subfigure}
    \begin{subfigure}{.33\textwidth}
      \centering
      \includegraphics[width=1\linewidth]{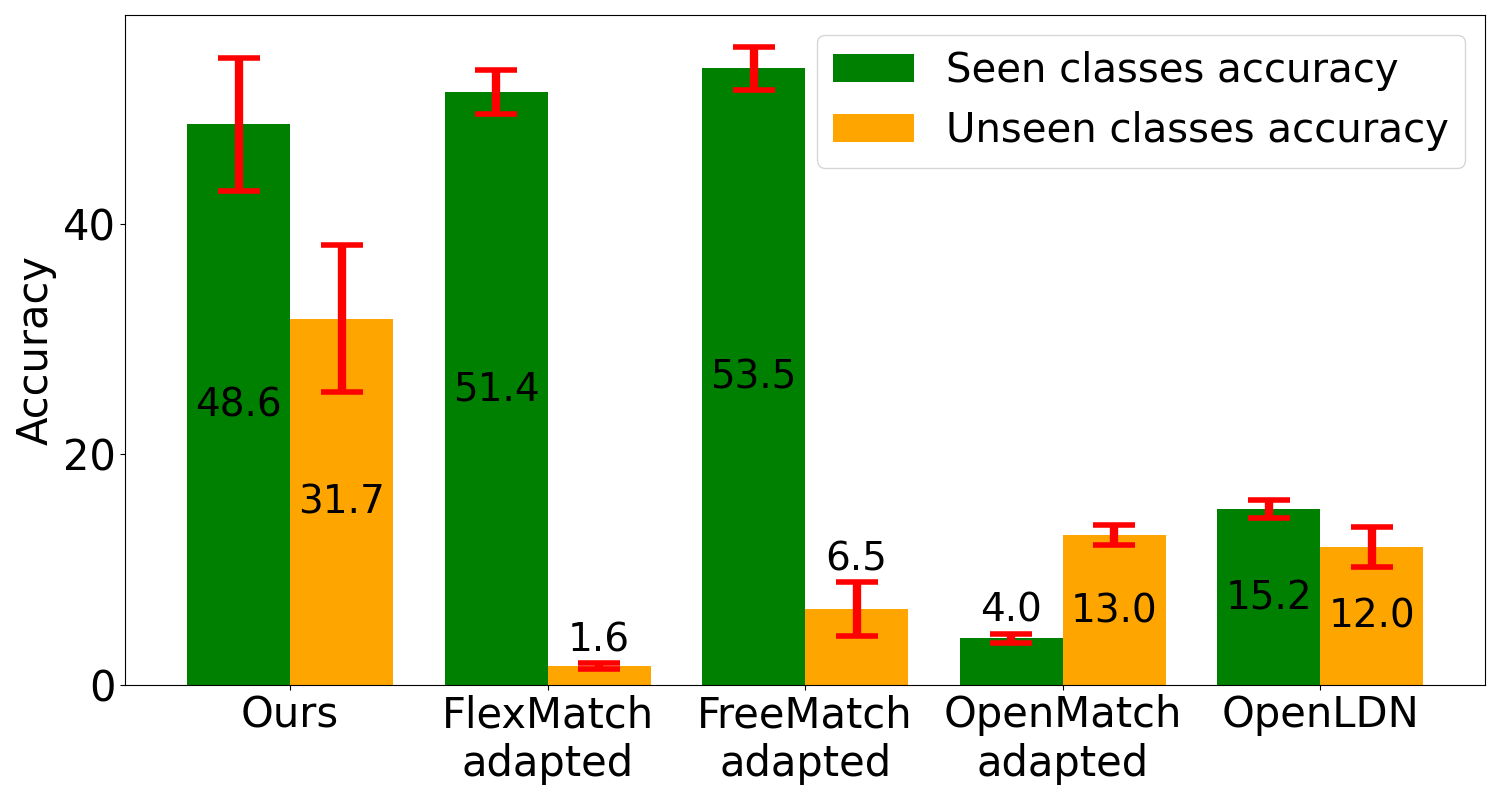}
      \caption{4 labels per class, 40 unseen classes}
      \label{subfig:4_labels_per_class_35_unseen}
    \end{subfigure}
    \begin{subfigure}{.33\textwidth}
      \centering
      \includegraphics[width=1\linewidth]{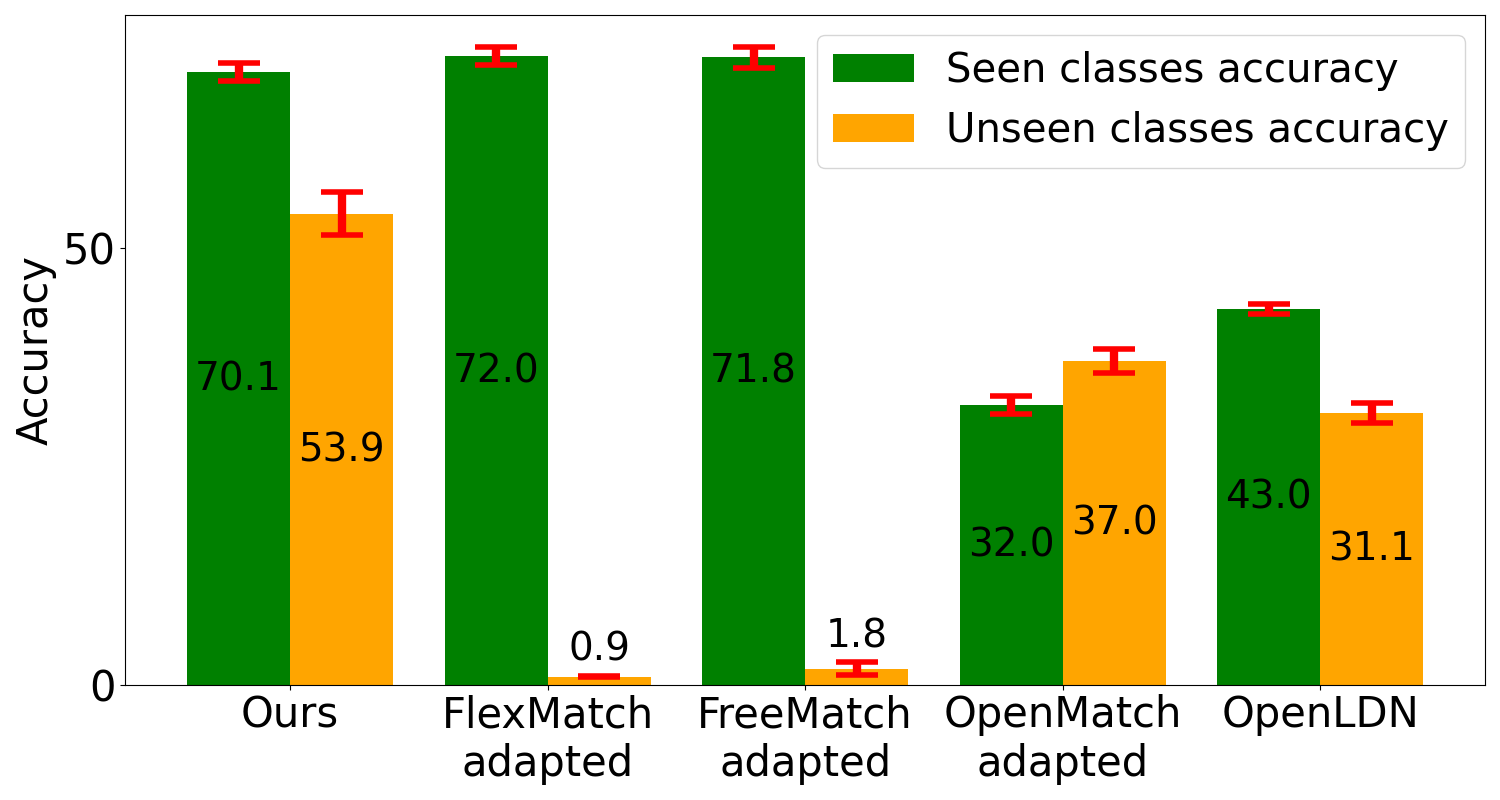}
      % \vspace{-0.62cm}
      \caption{25 labels per class, 40 unseen classes}
      \label{subfig:25_labels_per_class_35_unseen}
    \end{subfigure}
    
    \caption{Comparison of our method to the baselines discussed in Section~\ref{subsec:methodology}, case 1. All methods are trained on CIFAR-100, where (a),(b), and (c) differ in the size of the labeled set and the number of unseen classes. In the top row, we show the \emph{combined score} (\ref{eq:acc_combined}) of each method. In the bottom row, we show the \emph{seen accuracy} (\ref{eq:acc_seen}) and \emph{unseen accuracy} (\ref{eq:acc_unseen}) scores for each method. We clearly see that our method outperforms all baselines by significant margins. When compared to SSL methods - \freematch\  and \flexmatch,  we see that while the performance on the seen classes is comparable, our method improves the performance on the unseen classes quite drastically. Both \openmatch\  and \openldn\  are not suitable for the few-shot regime, and their performance is therefore poor.}
    \label{fig:1_main_results_different_data_sizes}
    \vspace{-0.42cm}
\end{figure*}

We start in Section~\ref{subsec:methodology} by describing the empirical settings used to evaluate our method in comparison to established baselines. In Section~\ref{sec:results} we report comprehensive results. In Section~\ref{sec:ablation} we report the results of an ablation study.

\subsection{Methodology and technical details}
\label{subsec:methodology}

\paragraph{Datasets.}
We use two benchmark image datasets in the experiments below: CIFAR-100 \citep{krizhevsky2009learning} and STL-10 \citep{coates2011analysis}. When using STL-10, we omitted the unlabeled split due to its inclusion of out-of-distribution examples.

\paragraph{Baselines.}
We compare our method to the following baselines: (i) SSL - \freematch\  \citep{DBLP:conf/iclr/Wang0HHFW0SSRS023} which performs well in the few-shot regime, and \flexmatch\  \citep{DBLP:conf/nips/ZhangWHWWOS21} which is the backbone of our own method. (ii) Open-set SSL - \openmatch\  \citep{DBLP:journals/corr/abs-2105-14148}. (iii) Open-world SSL: \openldn\  \citep{rizve2022openldn}, which includes its own adaptation to our scenario using \mixmatch\  \citep{berthelot2019mixmatch}. For \flexmatch\  and \freematch\  we employed the SSL evaluation framework established by \citep{usb2022}, ensuring a fair comparison. For \openmatch\  and \openldn\  we used the source code provided with the original papers. The methods were adapted to our scenario as detailed in Section~\ref{sec:adapt}.

\paragraph{Architectures and Hyper-parameters.}
When training \flexmatch, \freematch, and \openmatch, we employed Wide-ResNet-28 (WRN) \citep{zagoruyko2016wide}, trained with stochastic gradient descent optimizer, 64 batch size, 0.03 learning rate, 0.9 momentum and 5e-4 weight decay. In Fig~\ref{fig:1_main_results_different_data_sizes} we used 8 width factor, as was done in the original papers. Due to its heavy computational cost, in all other experiments, we reduced the width factor to 2. We note that the qualitative results remained the same.

When training \openldn, we used the official code, composed of a basic model and a \mixmatch\  model. Both models use ResNet-18, trained with Adam optimzier. The base model uses a 200 batch size and a 5e-4 learning rate. 
\mixmatch\  uses 64 batch size and 0.002 learning rate. 

The $\lambda$ for $\ell_{kl}$ was set to 1.5 in all experiments. We used NVIDIA GeForce RTX 2080 GPUs for all experiments.

\paragraph{Split of classes to seen and unseen.}

When training on CIFAR-100, the split of the 100 classes into seen and unseen classes was done randomly, repeating the exact same partition for all models. It's important to note that in the empirical evaluation of \openmatch, outlined in \citep{DBLP:journals/corr/abs-2105-14148}, the choice of unseen classes was determined by their super-class membership, introducing a potential bias. This may explain the differences in accuracy between the results we report below for \openmatch, and those reported in \citep{DBLP:journals/corr/abs-2105-14148}. To validate this point, we used their split method in a subset of our experiments, those involving the many-shot scenarios with 100 or 250 examples per class, replicating the reported results (see case 4 below).

\subsection{Results}
\label{sec:results}

\myparagraph{Case 1: Balanced labeled set, few-shot for seen classes.} 
We begin by comparing the performance of our method, using \flexmatch\ as the backbone, with two SSL methods -- \flexmatch\  and \freematch, one OSSL method -- \openmatch, and one OWSSL method -- \openldn. \flexmatch, \freematch, and \openmatch\  are adapted to our scenario, as described in Section~\ref{sec:adapt}. All methods are trained on CIFAR-100, with 35/40 unseen classes (similar qualitative results are obtained when the number of classes is varied, see case 5). The number of labeled points in each seen class varies, from one-shot with 1 example, very few-shot with 4 examples and few-shot with 25 labels, see Figs.~\ref{subfig:1_labels_per_class_35_unseen},\ref{subfig:4_labels_per_class_35_unseen},\ref{subfig:25_labels_per_class_35_unseen} respectively. Additional results with 35 unseen classes are shown in the \app~Fig.~\ref{fig:2_labels_per_class_35_unseen_wrn_28_2_combine}.

Inspecting the \emph{combined accuracy} score (\ref{eq:acc_combined}) in the top row of Fig.~\ref{fig:1_main_results_different_data_sizes}, we observe that our method outperforms the other baselines by large margins. In each case, we also plot the \textit{seen} and \textit{unseen} accuracy scores, (\ref{eq:acc_seen}) and   (\ref{eq:acc_unseen}) respectively. Evidently, the improvement over other SSL algorithms is caused primarily by their poor performance on unseen classes, which in turn is caused by their overconfidence in their erroneous classification of seen classes. This is a well-known problem \citep{nguyen2015deep} when attempting to add a reject option to deep models, which can be further appreciated from the comparison in the ablation study, Fig.~\ref{fig:Combine_score_kmeans_compare}.

\begin{figure*}[htb!]
\centering

        \begin{subfigure}{.33\textwidth}
    \centering
      \includegraphics[width=1\linewidth]{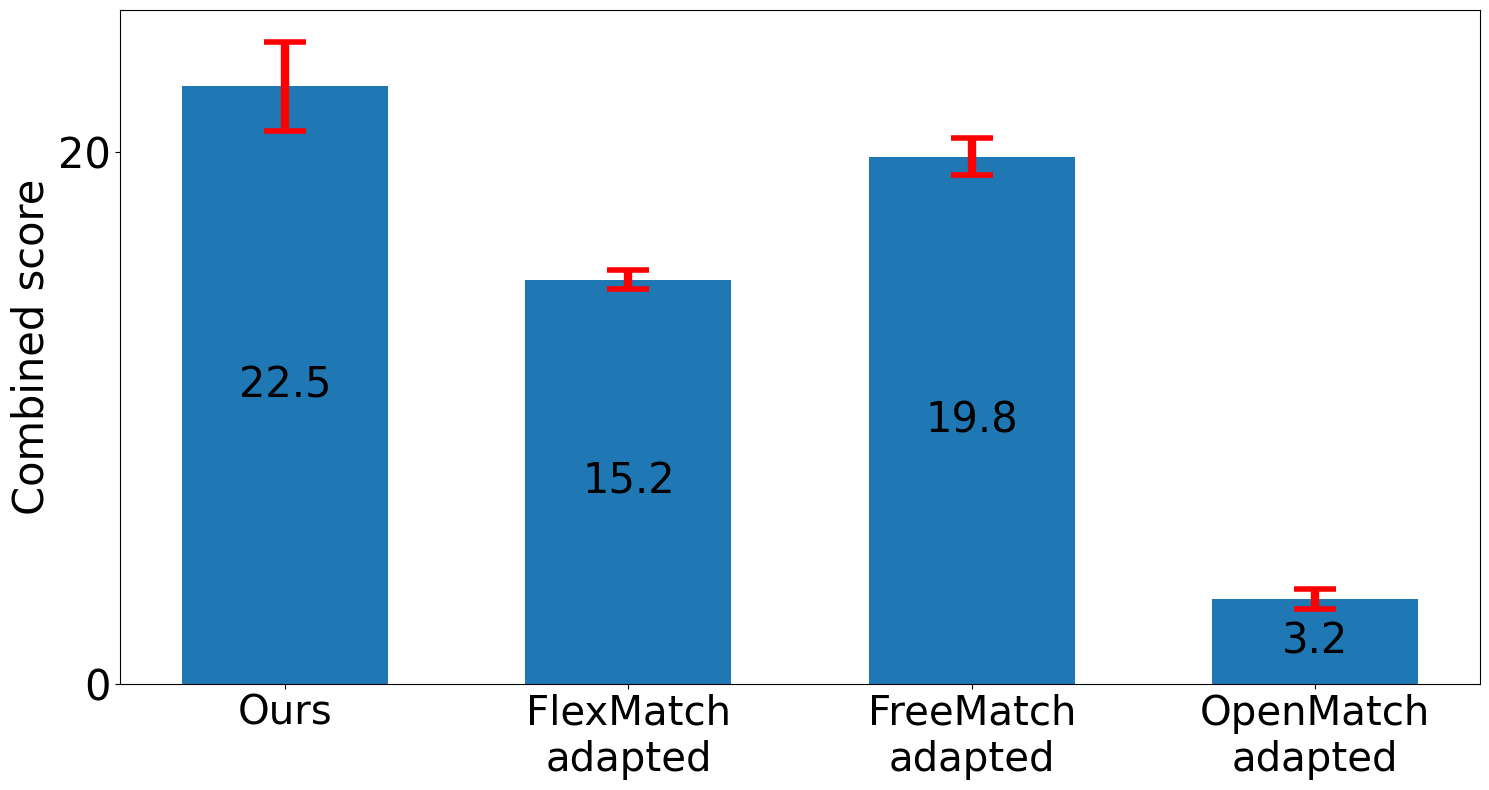}
    \end{subfigure}
    \begin{subfigure}{.33\textwidth}
    \centering
      \includegraphics[width=1\linewidth]{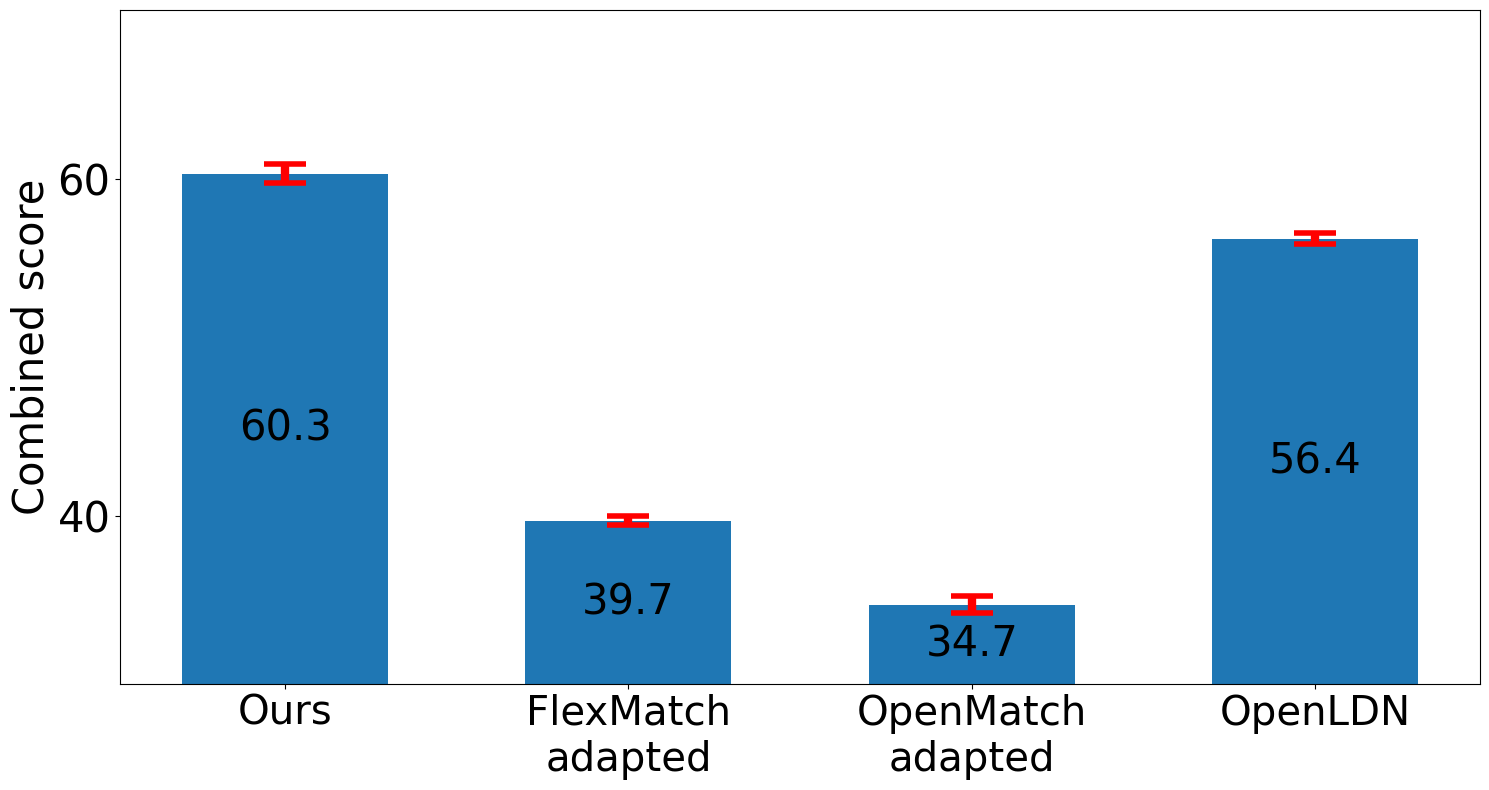}
    \end{subfigure}
    \begin{subfigure}{.33\textwidth}
    \centering
      \includegraphics[width=1\linewidth]{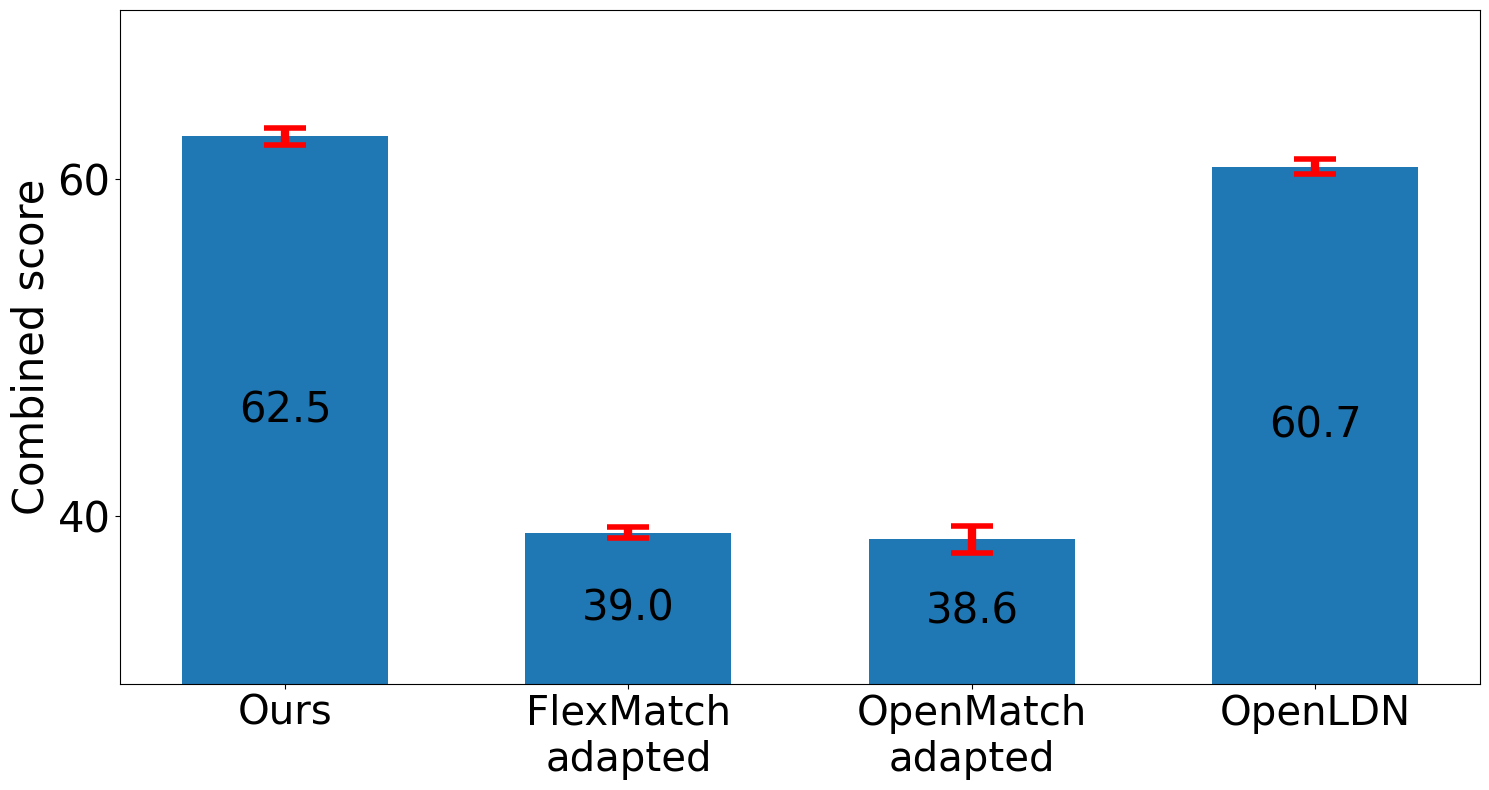}
    \end{subfigure}
    
    \begin{subfigure}{.33\textwidth}
      \centering
      \includegraphics[width=1\linewidth]{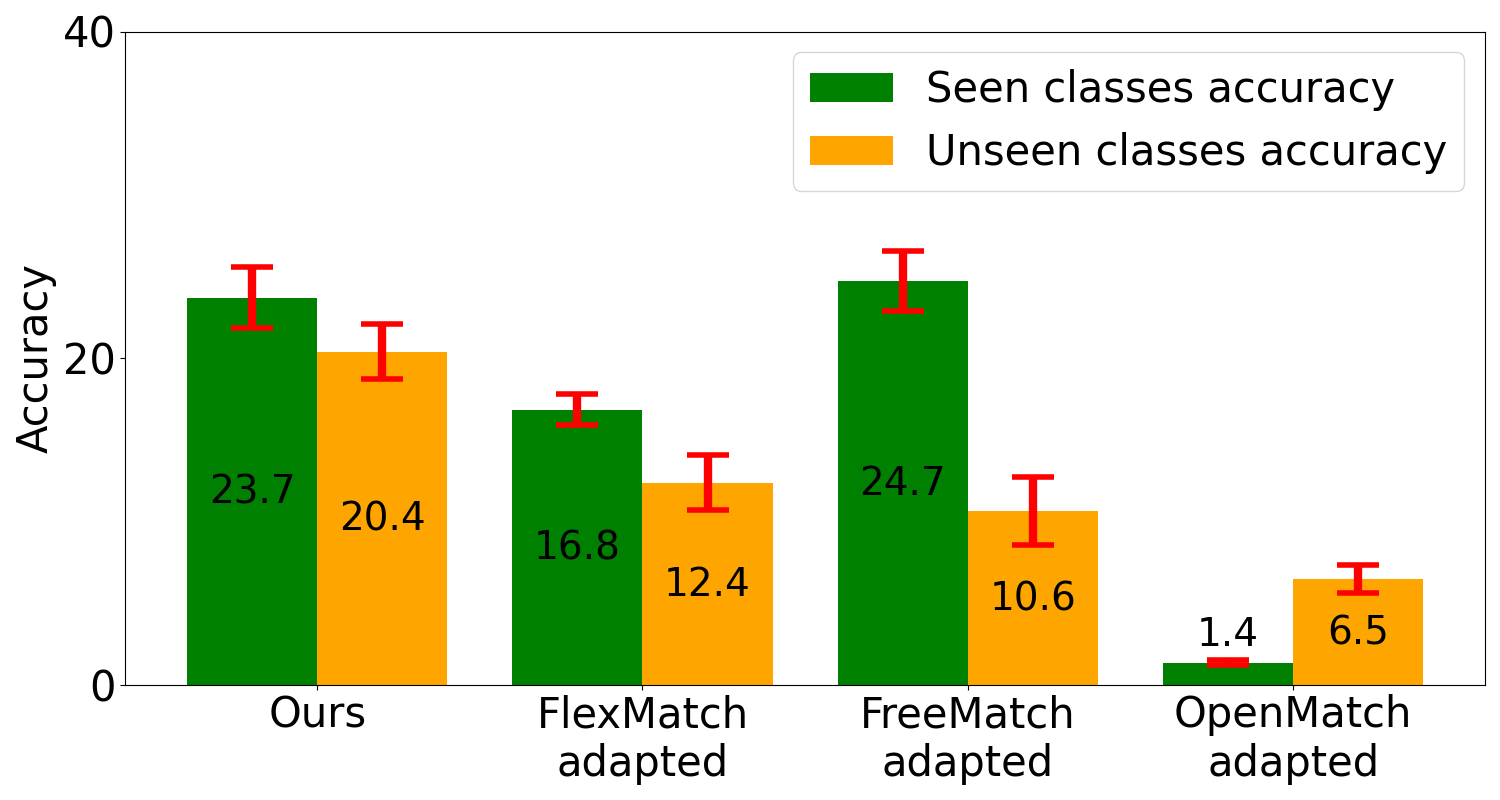}
      \caption{Case 2, total of 100 labeled examples}
      \label{fig:100_labels_random_set_wrn_28_2}
    \end{subfigure}
    \begin{subfigure}{.33\textwidth}
      \centering
      \includegraphics[width=1\linewidth]{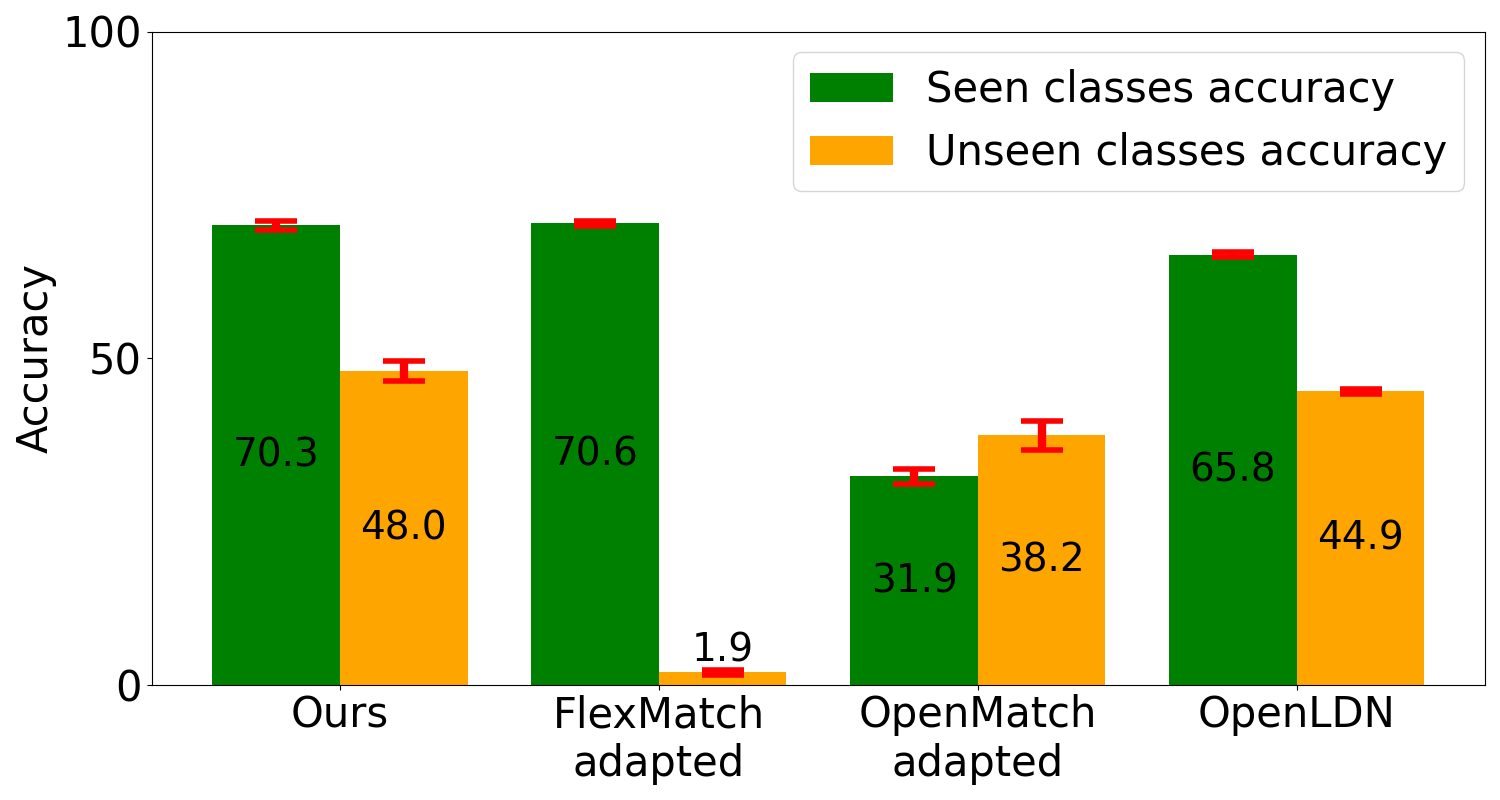}
      \caption{Case 4, 100 labels per class, 45 unseen classes}
      \label{fig:100_labels_per_class}
    \end{subfigure}
    \begin{subfigure}{.33\textwidth}
      \centering
      \includegraphics[width=1\linewidth]{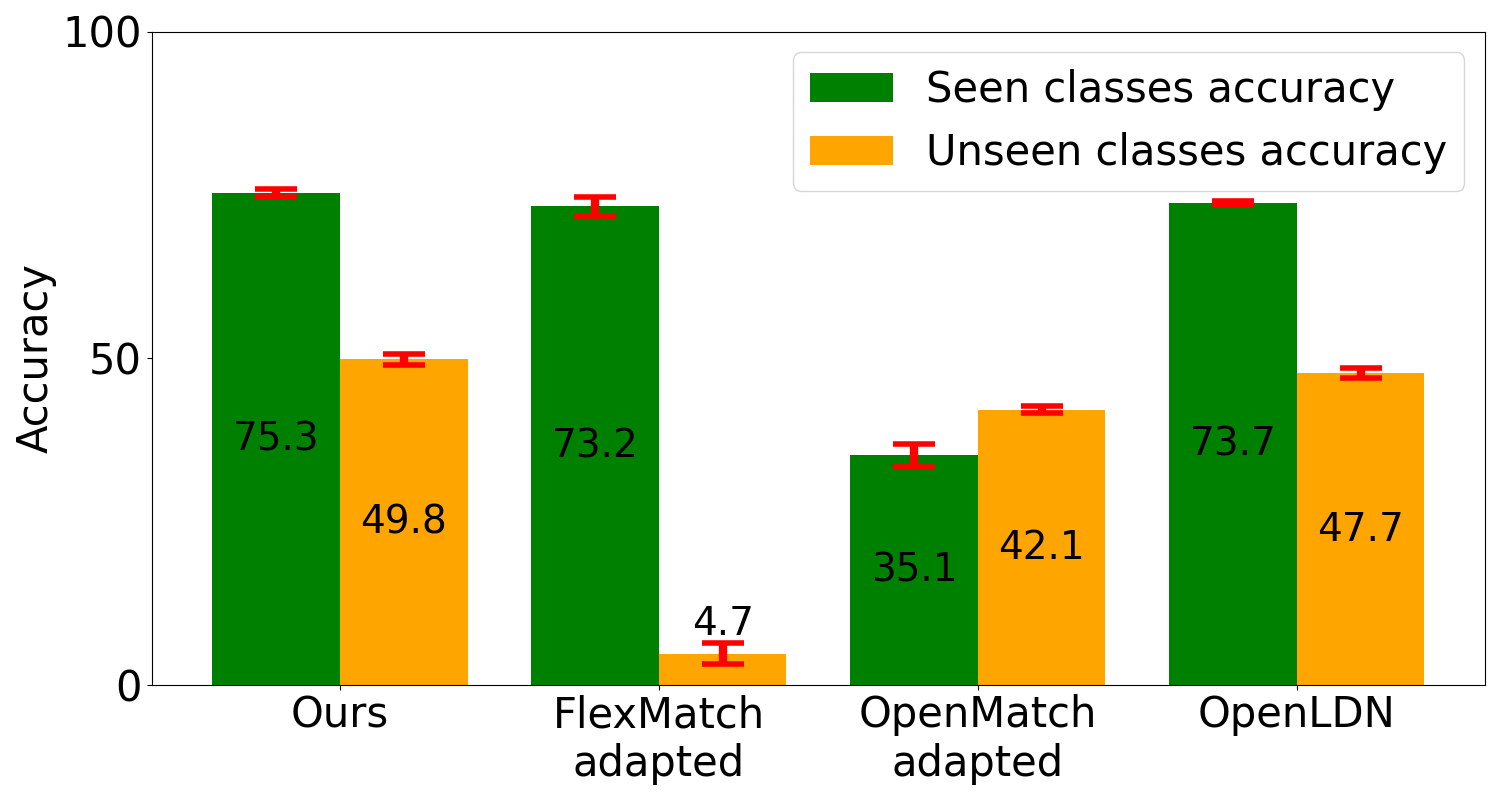}
      \caption{Case 4, 250 labels per class, 50 unseen classes}
      \label{fig:250_labels_per_class}
    \end{subfigure}
    
    % \vspace{-0.42cm}
    \caption{Additional results with CIFAR-100. (a) Case 2: total of 100 labeled examples sampled at random (see the caption of Fig.~\ref{fig:1_main_results_different_data_sizes}). On average, there are $36.61\pm1.38$ unseen classes, and $1.58\pm0.03$ labeled examples per seen class. (b-c) Case 4: Split is based on super-classes as explained in Section~\ref{subsec:methodology}. }
    
\end{figure*}

\begin{figure*}[htb!]
\centering

    \begin{subfigure}{.18\textwidth}
    \centering
     \includegraphics[width=1\linewidth]{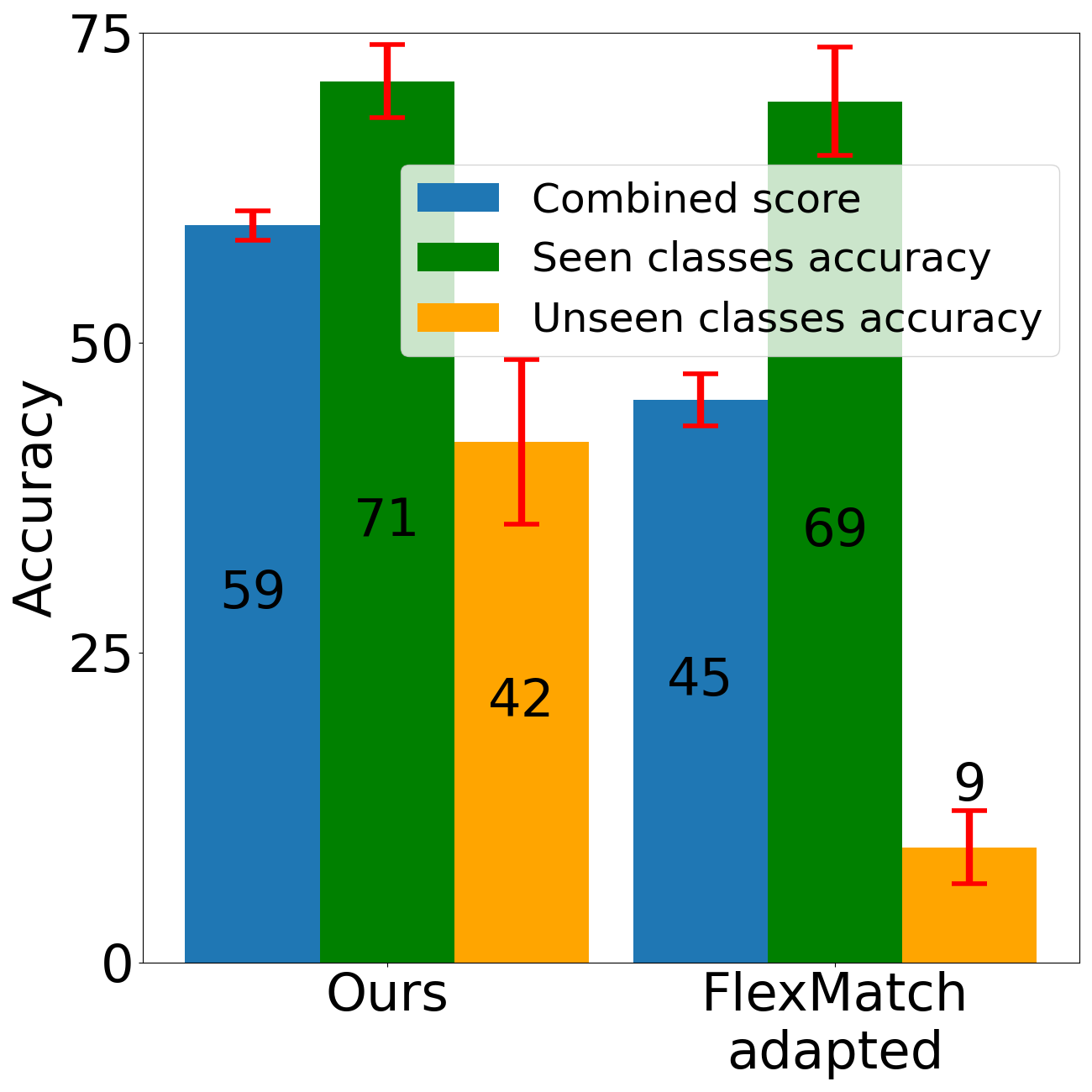}
      \caption{STL10, 40 labels per class, 4 unseen classes.}
    \label{fig:stl10_4_labels_per_class_4_unseen_wrn_28_2}
    \end{subfigure}
\hfill
\begin{subfigure}{.18\textwidth}
    \centering
      \includegraphics[width=1\linewidth]{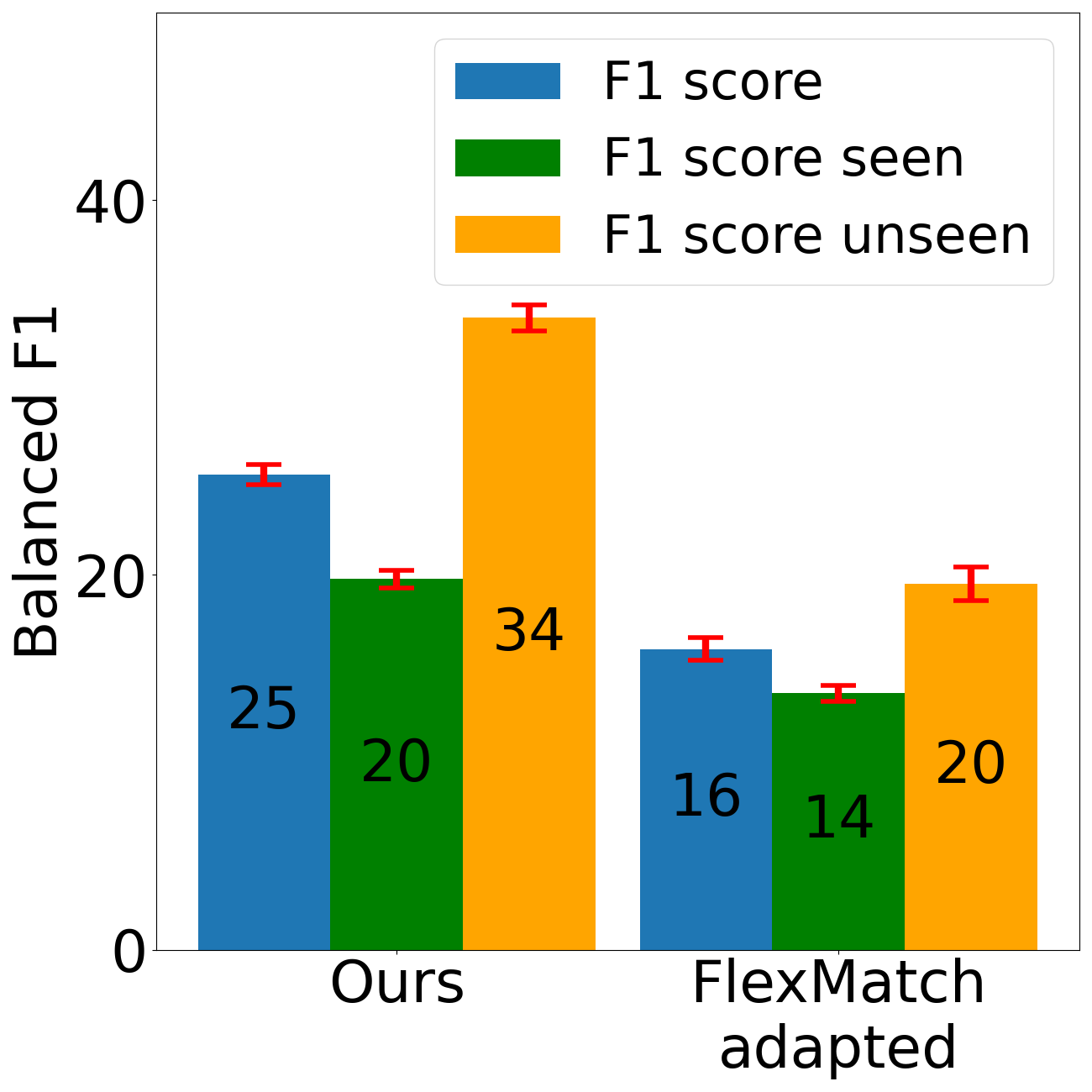}
    \caption{Modest long-tail, 165 labeled examples.}
    \label{fig:long_tail_lt_ratio_10_total_sum_165}
    \end{subfigure}
\hfill
    \begin{subfigure}{.18\textwidth}
    \centering
      \includegraphics[width=1\linewidth]{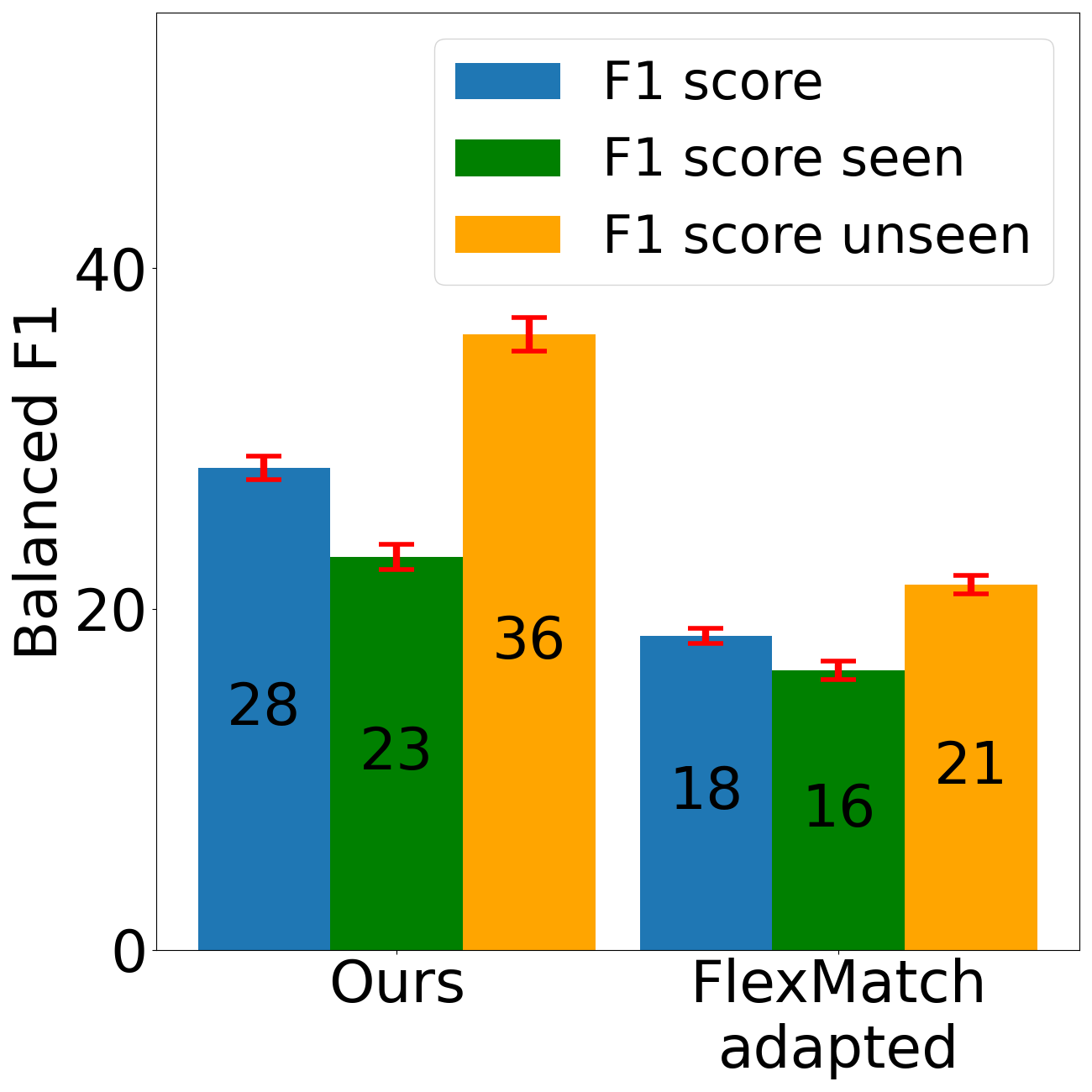}
    \caption{Modest long-tail, 260 labeled examples.}
    \label{fig:long_tail_lt_ratio_10_total_sum_260}
    \end{subfigure}
\hfill
    \begin{subfigure}{.18\textwidth}
    \centering
      \includegraphics[width=1\linewidth]{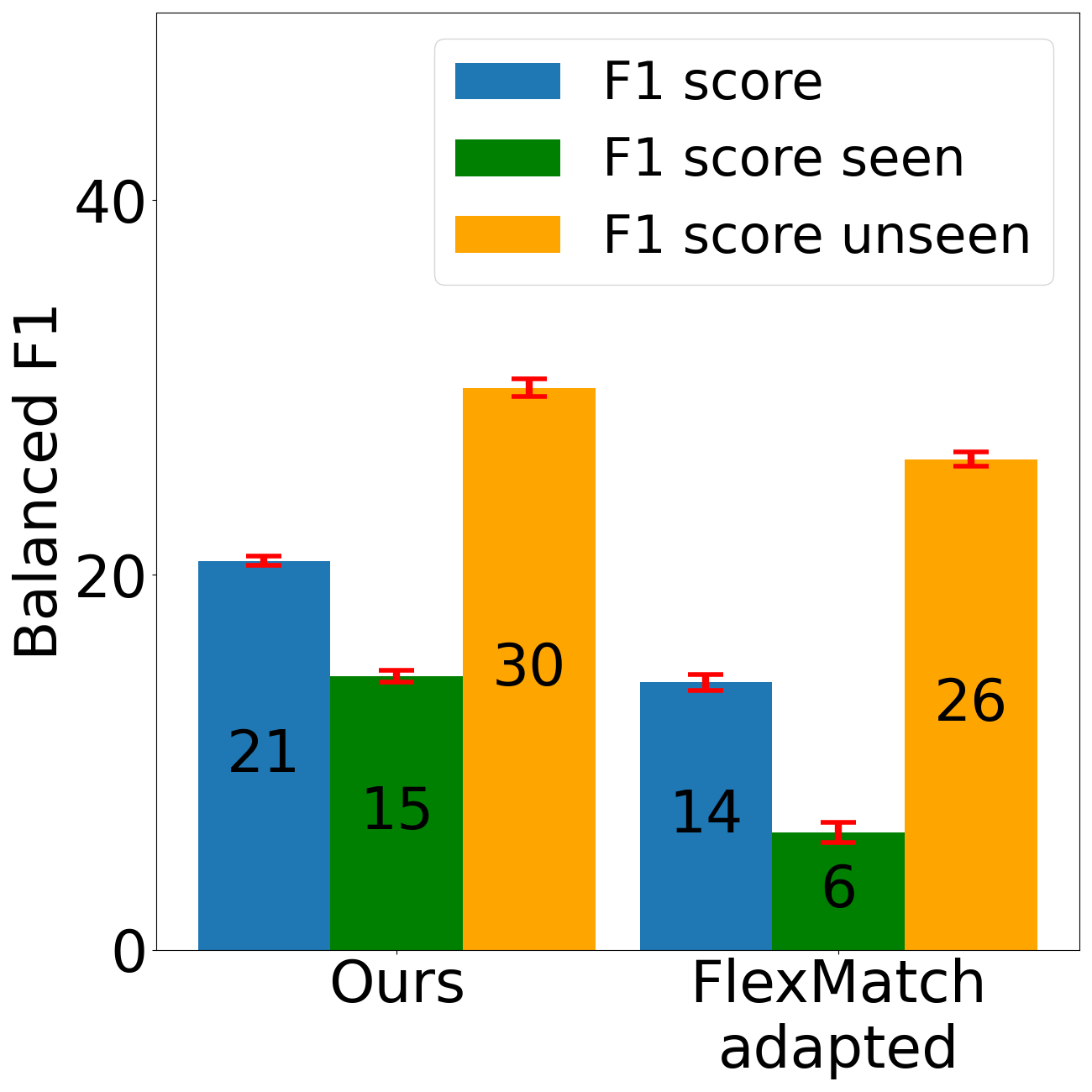}
    \caption{Sharp long-tail, 238 labeled examples.}
    \label{fig:long_tail_lt_ratio_50_total_sum_238}
    \end{subfigure}
\hfill
    \begin{subfigure}{.18\textwidth}
    \centering
      \includegraphics[width=1\linewidth]{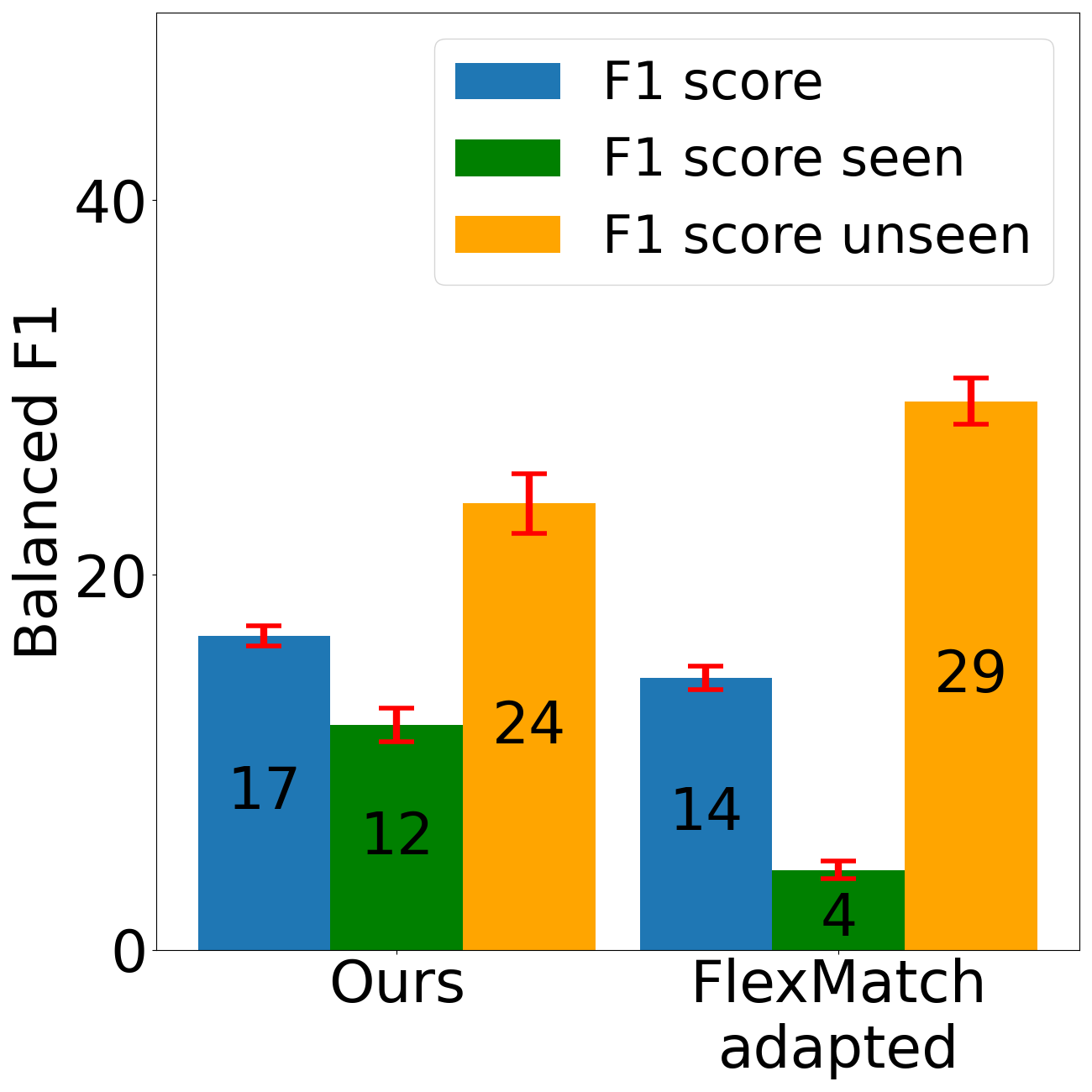}
    \caption{Sharp long-tail, 247 labeled examples.}
    \label{fig:long_tail_lt_ratio_100_total_sum_247}
    \end{subfigure}

% \hfill
% \begin{subfigure}{.15\textwidth}
%     \centering
%       \includegraphics[width=1\linewidth]{graphs/long_tail/lt_ratio_10/total_sum_165/f1_score_summarize_choose_unseen_count_by_test_1.png}
%     \caption{Modest long-tail, 165 labeled examples.}
%     \label{fig:long_tail_lt_ratio_10_total_sum_165}
%     \end{subfigure}
% \hfill
%     \begin{subfigure}{.15\textwidth}
%     \centering
%       \includegraphics[width=1\linewidth]{graphs/long_tail/lt_ratio_10/total_sum_260/f1_score_summarize_choose_unseen_count_by_test_1.png}
%     \caption{Modest long-tail, 260 labeled examples.}
%     \label{fig:long_tail_lt_ratio_10_total_sum_260}
%     \end{subfigure}
% \hfill
%     \begin{subfigure}{.15\textwidth}
%     \centering
%       \includegraphics[width=1\linewidth]{graphs/long_tail/lt_ratio_50/total_sum_238/f1_score_summarize_choose_unseen_count_by_test_1.png}
%     \caption{Sharp long-tail, 238 labeled examples.}
%     \label{fig:long_tail_lt_ratio_50_total_sum_238}
%     \end{subfigure}
% \hfill
%     \begin{subfigure}{.15\textwidth}
%     \centering
%       \includegraphics[width=1\linewidth]{graphs/long_tail/lt_ratio_100/total_sum_247/f1_score_summarize_choose_unseen_count_by_test_1.png}
%     \caption{Sharp long-tail, 247 labeled examples.}
%     \label{fig:long_tail_lt_ratio_100_total_sum_247}
%     \end{subfigure}
    
    \caption{(a) Case 1, STL10. (b-e) Case 3, CIFAR100 with 40 unseen classes. Here, the total number of labeled examples vary because the class distribution is long-tail, and the number of labeled examples from each class is different. Three different conditions are shown, corresponding to (b-c) modest long-tail distribution, (d) sharp long-tail distribution, and (e) sharper long-term distribution (see \app~\S\ref{sec:long-tail}).}
    % \noam{(c),(e) not finished running}
    \label{fig:long_tail_all}
\end{figure*}

% \begin{figure}[t!]
%     \begin{subfigure}{\columnwidth}
%       \centering
%       \includegraphics[width=1\linewidth,height=4cm]{graphs/random_set/100_labels/wrn_28_2/Combine_score_summarize.png}
%       \vspace{-0.62cm}
%       \caption{\textit{Combined score}}
%       \vspace{0.2cm}
%       \label{subfig:100_labels_random_set_wrn_28_2_combine}
%     \end{subfigure}
%     \begin{subfigure}{\columnwidth}
%       \centering
%       \includegraphics[width=1\linewidth,height=4.5cm]{graphs/random_set/100_labels/wrn_28_2/seen_unseen.png}
%       \vspace{-0.62cm}
%       \caption{\textit{Seen} and \textit{unseen accuracy}}
%       \label{subfig:100_labels_random_set_wrn_28_2_seen_and_unseen}
%     \end{subfigure}
%       \vspace{-0.42cm}
%     \caption{Case 2: CIFAR-100, a total of 100 labeled examples sampled at random (see the caption of Fig.~\ref{fig:1_main_results_different_data_sizes}). On average, there are $36.61\pm1.38$ unseen classes, and $1.58\pm0.03$ labeled examples per seen class. }
%     \label{fig:100_labels_random_set_wrn_28_2}
%       \vspace{-0.65cm}
% \end{figure}

It's worth noting that the subpar performance of \openldn\  can be attributed to the intended focus on scenarios with considerably higher availability of labeled data, see Figs.~\ref{fig:100_labels_per_class}-\ref{fig:250_labels_per_class} (case 4).  \openmatch\  addresses the open-set scenario, where out-of-distribution data may occur. %Their under-performance likely stems from being tailored to suit different conditions.%, which are sometimes evaluated using different metrics as discussed in Section~\ref{sec:scores}.\noam{the problem is that adapted SSL always preform better than \openmatch, even in scenarios with considerably higher availability of labeled data. we can see better performance of \openmatch when we look at the OSSL scores. so I propose to change the paragraph to: It's worth noting that the subpar performance of \openldn\  can be primarily attributed to the intended focus on scenarios with considerably higher availability of labeled data. Similarly, in the context of \openmatch\ , besides the scarcity of labeled data, an additional contributing factor lies in its customization for markedly distinct performance metrics.}

The same pattern of results can be seen when using a second image dataset, STL-10. In Fig.~\ref{fig:stl10_4_labels_per_class_4_unseen_wrn_28_2} we report results with 40 labeled examples for each of 6 seen classes, and  4 unseen classes. Since \openmatch\  and \openldn\  are not natively adapted to STL-10, we only compare our method, which is based on \flexmatch, to plain \flexmatch\  adapted to our scenario. 

\myparagraph{Case 2: Non-uniform labeled set, approximately one-shot.} 
We now move on to address the more general scenario as envisioned above, where the labels are obtained as follows: To begin with, the learner is given a large set of unlabeled data randomly collected. Additionally, the learner is restricted to the labeling of only $n$ randomly chosen examples from the unlabeled set, thus establishing the labeled training set. Typically, the labeled set is not class-balanced. In our experiments with $n=100$, the actual number of examples per seen class varied from 1 to 4. The remaining examples compose the unlabeled training set.

Results are shown in Fig.~\ref{fig:100_labels_random_set_wrn_28_2}, for CIFAR-100 and $n=100$ labeled examples. Once again, when inspecting the \emph{combined score} (\ref{eq:acc_combined}) shown in the top row, our method outperforms all other baselines. Given the technical difficulties that this unusual setup poses, we could not obtain reliable results with \openldn. Nevertheless, we expect \openldn\  to perform poorly in these conditions, similarly to the case of 1 labeled example per seen class (see Fig.~\ref{subfig:1_labels_per_class_35_unseen}).

\paragraph{Case 3: Non-uniform class frequencies.} 

Our method, as described in Section~\ref{sec:ourmethod}, is able to handle non-uniform class distributions, where the loss component $l_{kl}$ defined in (\ref{eq:l_kl}) is designed to reduce the discrepancy between the empirical class frequencies and the prior class distribution. To evaluate the effectiveness of our method in this regard, we now test our method while using dataset with imbalanced class distribution. Specifically, we use a few variants of the long-tail CIFAR100 dataset, described in \citep{cao2019learning}. 

When evaluating the outcome of classification on imbalanced datasets, class accuracy is a very skewed measure of performance. As customary, we replace the combined score (\ref{eq:acc_combined}) by a score recommended for use when using imbalanced datasets. Specifically, we use the multi-class balanced F1-score \citep{grandini2020metrics}, computed as customary while using the following multi-class balanced precision and recall scores:
\begin{itemize}
    \item \textit{Multi-class balanced precision}
    \begin{equation*}
    \mbox{Balanced\_Precision} = \frac{1}{|C|}\sum_{k\in C} \frac{TP_k}{TP_k + FP_k} 
    \end{equation*}
    \item \textit{Multi-class balanced recall}
    \begin{equation*}
    \mbox{Balanced\_Recall} = \frac{1}{|C|}\sum_{k\in C} \frac{TP_k}{TP_k + FN_k} 
    \end{equation*}
    \end{itemize}
Above, $TP_k,FP_k,FN_k$ denote respectively the True Positive score of class $k$, its False Positive score, and its False Negative score.  

Results - balanced F1 scores of the different methods, are shown in Figs.~\ref{fig:long_tail_lt_ratio_10_total_sum_165}-\ref{fig:long_tail_lt_ratio_10_total_sum_260} for modest long-tail imbalance, and Figs.~\ref{fig:long_tail_lt_ratio_50_total_sum_238}-\ref{fig:long_tail_lt_ratio_100_total_sum_247} for large long-tail imbalance. Clearly the method is able to handle these challenging scenarios, significantly improving the outcome of plain \flexmatch.  
%\noam{need to add a sentence explaining the results with the unseen score.}
%\noam{do we need to explain what is "F1 score seen\\unseen"? or is it obvious?}
Note that in these conditions, where almost all the points in the unlabeled set belong to unseen classes (see \app~\S\ref{sec:long-tail}), both methods predict unseen classes better than seen classes. 

\myparagraph{Case 4: Many-shot zero-shot SSL.} 
In our final setting, we increase the number of labeled examples per class, thus obtaining scenarios that are more favorable to the alternative methods, especially \openmatch\  and \openldn.  Results\footnote{Since \flexmatch\  and \freematch\  perform similarly in the many-shot case, we only show results using \flexmatch. } with 5 repetitions
are shown in Figs.~\ref{fig:100_labels_per_class}-\ref{fig:250_labels_per_class}. %{fig:2500_labels_per_class_50_unseen_wrn_28_2} (see also Fig.~\ref{fig:100_labels_per_class_45_unseen_wrn_28_2} in appendix). 
For completeness, we report additional scores, which are sometimes favored when evaluating open-set SSL methods, in the \app~Fig.~\ref{fig:100_and_250_labels_per_class_OSSL_score}.
Our method significantly outperforms all baselines, though the benefit against \openldn\  decreases.

\begin{figure}[htb!]
    % \begin{subfigure}{\columnwidth}
    %   \centering
    %   \includegraphics[width=1\linewidth]{graphs/25_per_class/10_unseen/wrn_28_8/Combine_score_summarize_withoutspace.png}
    %   % \vspace{-0.62cm}
    %   % \caption{\textit{Combined score}}
    %   % \vspace{0.2cm}
    %   % %\label{subfig:25_labels_per_class_10_unseen_wrn_28_8_combine}
    %   \caption{\textit{Combined score}}
    % \end{subfigure}
    % \begin{subfigure}{\columnwidth}
    %   \centering
    %   \includegraphics[width=1\linewidth]{graphs/25_per_class/10_unseen/wrn_28_8/seen_unseen.png}
    %   % \vspace{-0.62cm}
    %   % \caption{\textit{Seen} and \textit{unseen accuracy}}
    %   % \label{subfig:25_labels_per_class_10_unseen_wrn_28_8_seen_and_unseen}
    %   \caption{\textit{Seen} and \textit{unseen accuracy}}
    % \end{subfigure}
    \begin{subfigure}{\columnwidth}
      \centering
      \includegraphics[width=1\linewidth]{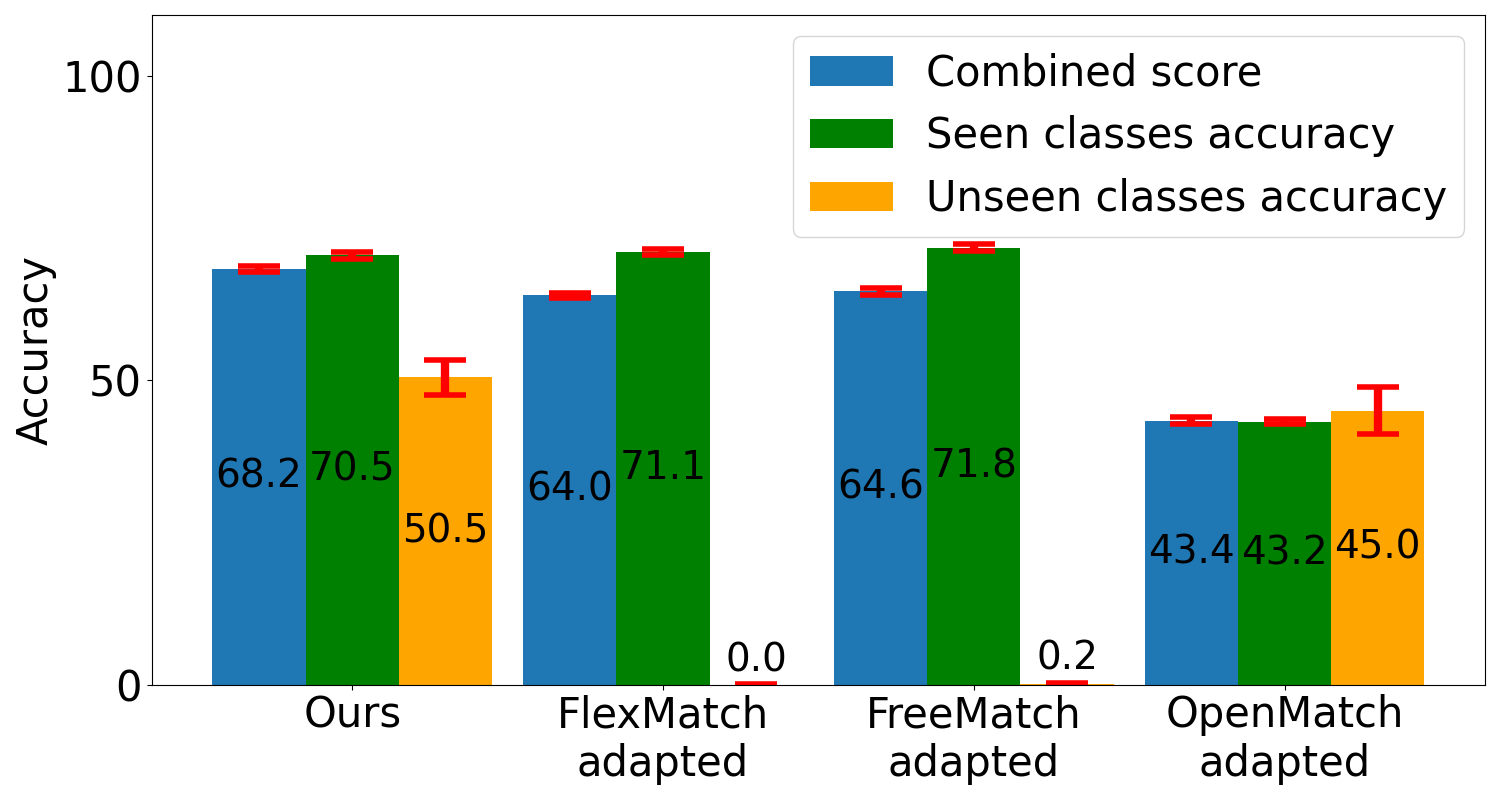}
      % \vspace{-0.62cm}
      % \caption{\textit{Seen} and \textit{unseen accuracy}}
      % \label{subfig:25_labels_per_class_10_unseen_wrn_28_8_seen_and_unseen}
%      \caption{\textit{Combine score}, \textit{Seen} and \textit{Unseen accuracy}}
    \end{subfigure}
    \caption{Case 1, CIFAR-100 with 25 labels per class and 10 unseen classes. While the gap between the methods is smaller as compared to Fig.~\ref{fig:1_main_results_different_data_sizes}, our method is still the best performer. }
    \label{fig:25_labels_per_class_10_unseen_wrn_28_8}
\vspace{-0.3cm}
\end{figure}

\myparagraph{Case 5: A few zero-shot classes.}
Given the analysis above, we expect that the relative advantage of our method will decrease as the number of unseen classes decreases. To investigate this point, we repeat the experiment whose results are reported in Fig.~\ref{subfig:25_labels_per_class_35_unseen}, while reducing the number of unseen classes from $35$ to $10$. Results are shown in Fig.~\ref{fig:25_labels_per_class_10_unseen_wrn_28_8}. While the advantage of our suggested method indeed decreases as expected, our method still outperforms all baselines by a large and significant margin. 

\subsection{Ablation study}
\label{sec:ablation}

\myparagraph{SSL adaptation.}
To adapt the SSL methods to our scenario, we introduced in Section~\ref{sec:adapt} a two-step adaptation process: (i) Points with low classification confidence are rejected from the classifier trained on seen classes. (ii) K-means clustering is used to partition the rejected points within the feature space. This partitioning is then matched to unseen classes. Without this adaptation, SSL methods would not identify anything as unseen, and would therefore erroneously match all points from unseen classes to some seen class. In each case, the rejection threshold was optimized to give each method its best possible rejection outcome.

\begin{figure}[htb!]
    \begin{subfigure}{0.47\columnwidth}
      \centering
      \includegraphics[width=1\linewidth]{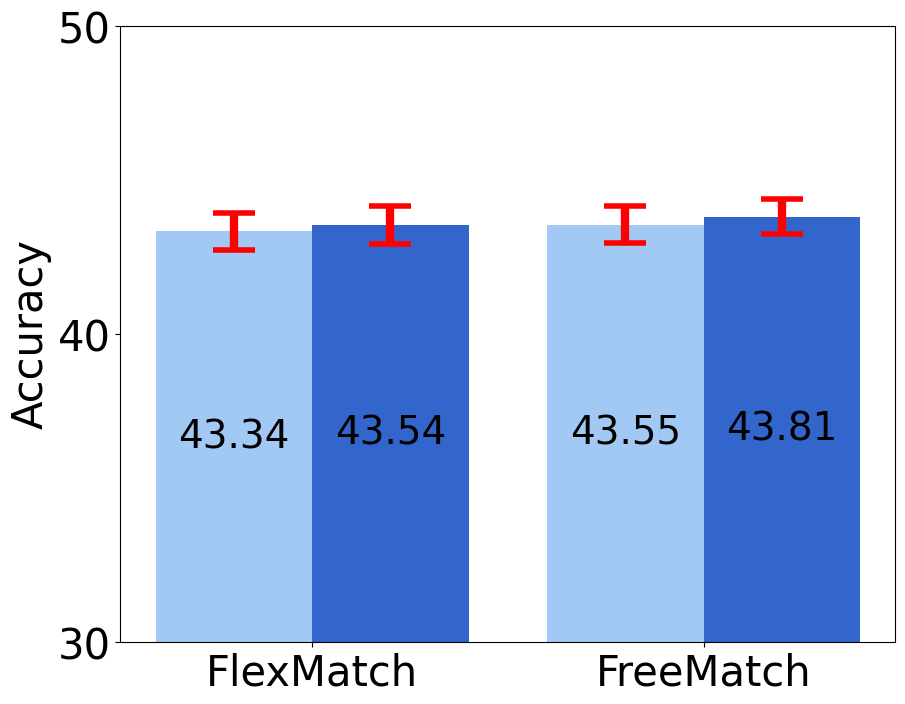}
      % \vspace{-0.62cm}
      \caption{\centering 25 labels per class, 40 unseen classes.}
      \label{subfig:25_labels_per_class_40_unseen_wrn_28_8_Combine_score_kmeans_compare}
    \end{subfigure}
    \hfill
    \begin{subfigure}{0.47\columnwidth}
      \centering
      \includegraphics[width=1\linewidth]{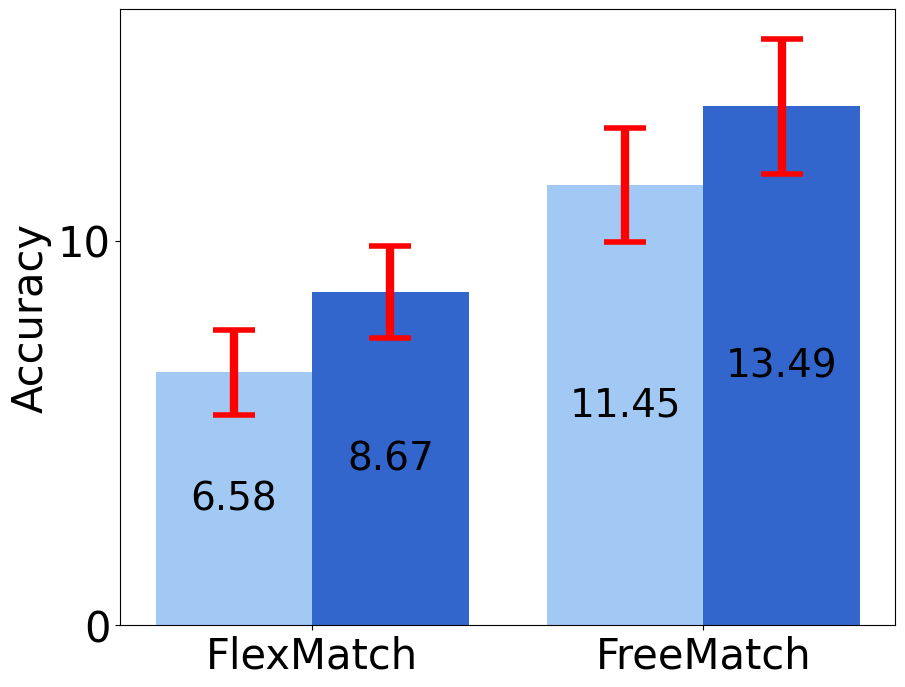}
      % \vspace{-0.62cm}
      \caption{\centering 1 label per class, 35 unseen classes.}
    \label{subfig:1_labels_per_class_35_unseen_wrn_28_2_Combine_score_kmeans_compare}
    \end{subfigure}
      \vspace{-0.2cm}
    \caption{\textit{Combined score} (\ref{eq:acc_combined}) of \flexmatch\  and \freematch\  with and without the being adapted to our scenario (see Section~\ref{sec:adapt}), on CIFAR-100. \emph{With adaptation} - dark blue, \emph{without adaptation} - light blue.}
    \label{fig:Combine_score_kmeans_compare}
     \vspace{-0.3cm}

\end{figure}

Here we evaluate the added value of this adaptation, as shown in Fig.~\ref{fig:Combine_score_kmeans_compare}. Evidently, the impact is minor (especially in the mid-shot regime). The reason is that the optimal threshold tends to be relatively low, resulting in a relatively modest count of predicted unseen instances. This count often pales in comparison to the actual prevalence of unseen classes within the test data. The underlying reason for this failure is the known property of deep models to be over-confident in their predictions. It is therefore hard to distinguish between seen and unseen classes based on the confidence threshold.

\myparagraph{Convergence time}
%\label{sec:convergence_time}
As described in Section~\ref{sec:ourmethod}, sometimes the SSL training, performed by our method, is stopped relatively early. This happens when the new loss term $l_{kl}$ suddenly declines sharply, and is usually correlated with a small number of labeled examples per seen class. As a result, it so happens that our model needs much fewer epochs to converge, as compared to the original model that serves as its backbone, \flexmatch. This result is shown in Table~\ref{tab:convergence_epoch}.
\begin{table}[ht]
% \vspace{0.2cm}
  \centering
  \begin{tabular}{ccc}
    
    Labels per & Unseen  & Convergence  \\
    class & classes & epoch \\
    \hline
    1           & 35         & $224.6 \pm 44.19$           \\
    4           & 40         & $523.0 \pm 24.61$           \\
    25           & 40         & $726.6 \pm 8.85$          
  \end{tabular}
  \caption{The mean number of epochs to convergence, in a scenario where \flexmatch\  needs 1024 epochs to converge. }
  \label{tab:convergence_epoch}
  % \vspace{-0.4cm}
\end{table}

\subsection{Summary and Discussion}
We investigated Semi-Supervised Learning (SSL) in a small sample framework with few-shot and zero-shot classes, thereby confronting a previously unexplored real-life challenge. To address this challenge, we add a penalty term -- the KL-divergence between the empirical and true class frequencies vectors -- to the loss of an established state-of-the-art SSL method, such as \flexmatch\ or \freematch. This enhancement mediates the transfer of knowledge from the few-shot to the zero-shot classes. Thus, the proposed approach allows us to convert any state-of-the-art SSL algorithm to readily face the new challenge.

Our extensive empirical evaluation provides strong evidence for the effectiveness of our approach in addressing the previously mentioned challenge when compared to alternative baselines. These baselines include SSL and open-set SSL methods (adapted to accommodate our scenario as detailed in Section \ref{sec:adapt}), along with open-world SSL. It is worth mentioning that our algorithm emerges as a top-performer by significant margins, particularly in situations characterized by the scarcity of labeled data.
%\paragraph{Future work}

In future work, we will investigate the dynamic incorporation of the penalty term throughout the training process. This adaptive integration holds promise for enhancing the robustness and versatility of our approach.

\clearpage
% \paragraph{Acknowledgement}
% This work was supported by grants from the Israeli Council of Higher Education and the Gatsby Charitable Foundations.

%\bibliography{aaai23}
{
    \small
    \bibliographystyle{ieeenat_fullname}
    \bibliography{aaai23}
}

\clearpage
%to fix in sty file: change to \def\copyright@on{T}

\appendix
\section*{Supplementary material}

% \section{Few labeled examples}
% In Fig.~\ref{fig:1_main_results_different_data_sizes}, we experimented with our suggested method using either $1$ or $4$ labels per class. These choices were made to show that the suggested method works where $\mathcal{L}$ is small, and the specific values were selected for better comparison to previous art. Here, we show that other sizes of $\mathcal{L}$ could be chosen. In Fig.~\ref{fig:2_labels_per_class_35_unseen_wrn_28_2_combine}, similarly to Fig.~\ref{fig:1_main_results_different_data_sizes}, we train WRN on CIFAR-100, changing the labeled set to contain either $2$ or $3$ labels per class. 

% \begin{figure}[htb!]
%     \begin{subfigure}{.4\columnwidth}
%       \centering
%       \includegraphics[width=1\linewidth]{graphs/2_per_class/35_unseen/wrn_28_2/Combine_score_summarize.png}
%       \vspace{-0.62cm}
%       \caption{2 labels per class}
% %      \vspace{0.2cm}
%       \label{subfig:2_labels_per_class_35_unseen_wrn_28_2_combine}
%     \end{subfigure}
%     \hfill
%     \begin{subfigure}{.4\columnwidth}
%       \centering
%       \includegraphics[width=1\linewidth]{graphs/3_per_class/35_unseen/wrn_28_2/Combine_score_summarize.png}
%       \vspace{-0.62cm}
%       \caption{3 labels per class}
% %      \vspace{0.2cm}
%       \label{subfig:3_labels_per_class_35_unseen_wrn_28_2_combine}
%     \end{subfigure}
%     \caption{\textit{Combined score}, CIFAR-100, 35 unseen classes.}
%     \label{fig:2_labels_per_class_35_unseen_wrn_28_2_combine}
%       \vspace{-0.3cm}
% \end{figure}

\section{Alternative Scores}

%\paragraph{Additional scores used in the OSSL literature.}
In OSSL, where the goal is only to reject unseen classes and not to classify them, different scores are sometimes used. For completeness, we report these scores below for cases favorable to traditional OSSL with many missing classes and sufficient labels for each seen class. 
%\subsubsection{open SSL measurements}
\begin{itemize}
    \item \textit{Closed accuracy}
    \begin{equation}
    \label{eq:closed_accuracy}
      \frac{1}{|\mathcal{T}_{seen}|}\sum_{x_i\in \mathcal{T}_{seen}}\mathds{1}_{[f|_{seen}(x_i)=y_i]}  
    \end{equation}
    Here, $f|_{seen}$ is the prediction of $f$ when forced to choose a class out of $C_{seen}$. Thus an OSSL method is not allowed to reject. Other methods are not allowed to classify points for labels in $C_{unseen}$.
    \item \textit{Unknown\ accuracy}\\
    \begin{equation}
    \label{eq:unknown_accuracy}
      \frac{1}{|\mathcal{T}_{unseen}|}\sum_{x_i\in \mathcal{T}_{unseen}}\mathds{1}_{[f(x_i)\in C_{unseen}]}  
    \end{equation}
This score captures the rejection accuracy.

    \item \textit{AUROC} 
    \begin{equation}
    \label{eq:AUROC}
      Area~under~ROC~curve~of~'reject'~classifier
    \end{equation}
\end{itemize}

Note the difference between \textit{seen classes accuracy} and \textit{closed accuracy}: the first penalizes for \textit{seen} points recognized as \textit{unseen}, while the second doesn't, and therefore \textit{seen classes accuracy}  is always lower than \textit{closed accuracy}.  

\begin{figure}[htb!]
\centering
    \begin{subfigure}{.33\textwidth}
    \centering
      \includegraphics[width=1\linewidth]{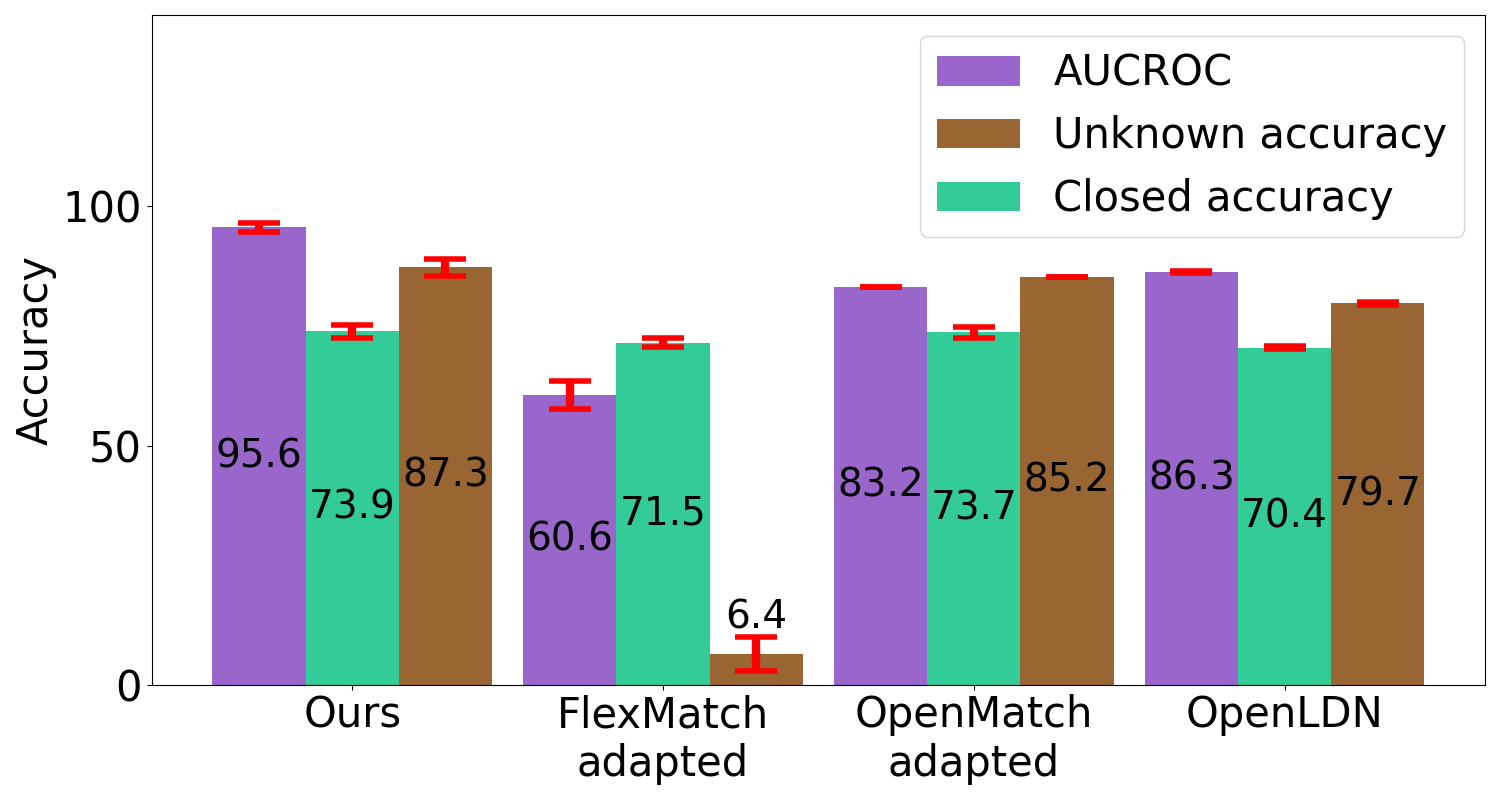}
      \caption{Case 4, 100 labels per class, 45 unseen classes}
    \end{subfigure}
    
    \begin{subfigure}{.33\textwidth}
      \centering
      \includegraphics[width=1\linewidth]{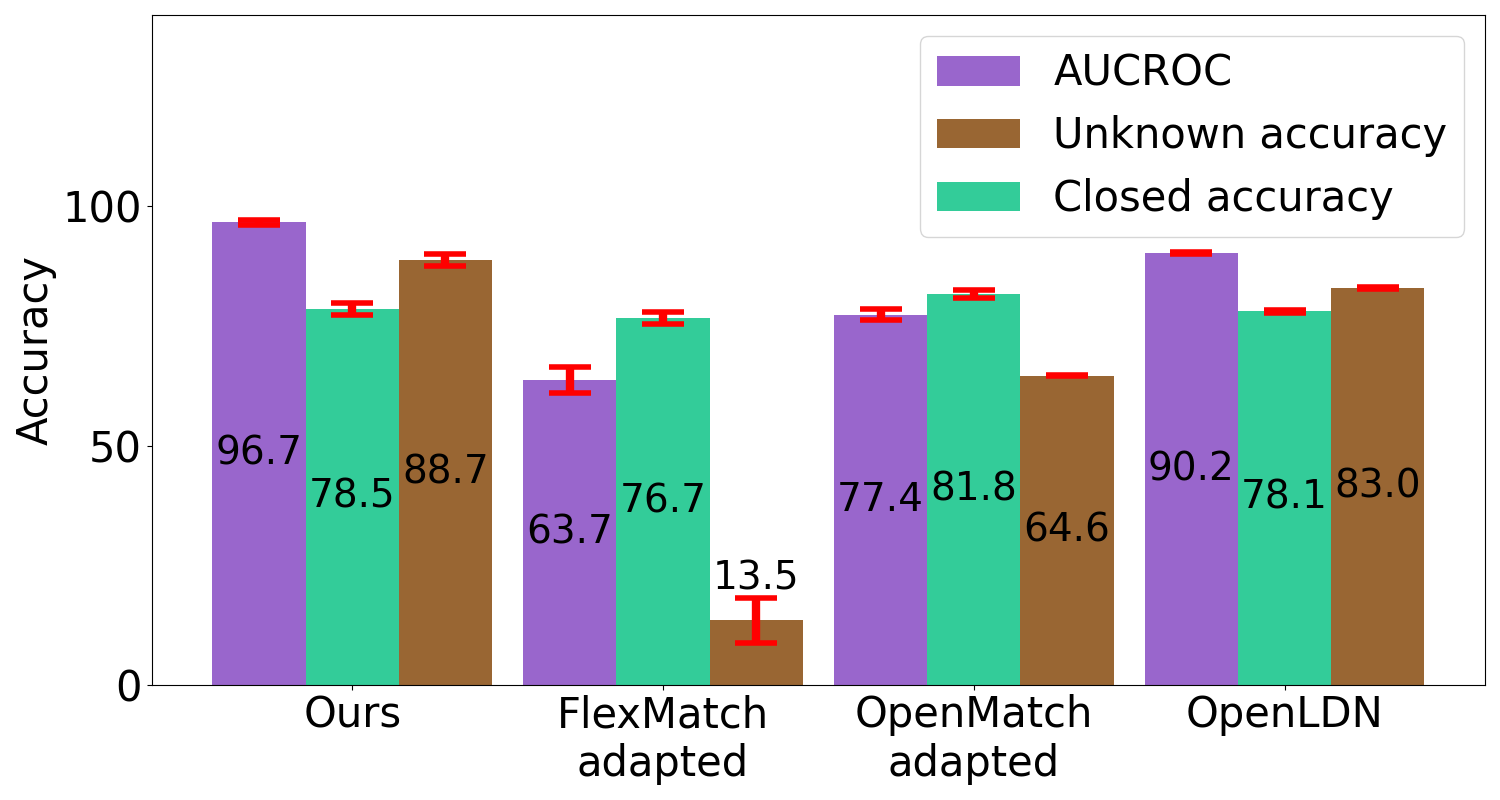}
      \caption{Case 4, 250 labels per class, 50 unseen classes}
    \end{subfigure}
    
    \caption{Alternative scores corresponding to the results shown in Figs.~\ref{fig:100_labels_per_class}-\ref{fig:250_labels_per_class}, using OSSL scores (\ref{eq:closed_accuracy})-(\ref{eq:AUROC}) as described in Section~\ref{sec:scores}.}
    \label{fig:100_and_250_labels_per_class_OSSL_score}
%    \vspace{-0.6cm}
\end{figure}

% \section{Extending traditional SSL for long tail scenario}
% In the case of the long tail scenario, we need to use another method to determain the \daphna{I have briefly explained it in case 3 (line t52), drop it}


\section{Additional Results}

\begin{figure}[htb]
\begin{center}
    \begin{subfigure}{.3\columnwidth}
      \centering
      \includegraphics[width=1\linewidth]{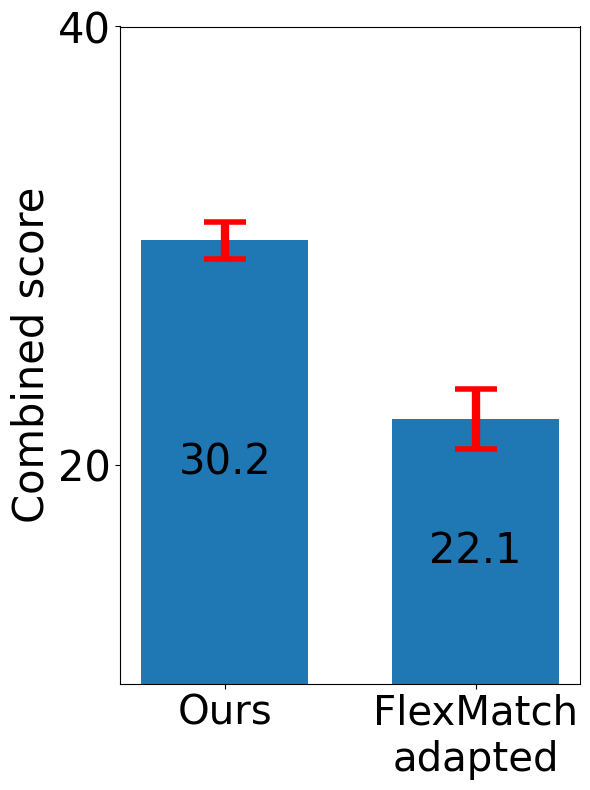}
      \vspace{-0.62cm}
      \caption{2 labels per class}
%      \vspace{0.2cm}
      \label{subfig:2_labels_per_class_35_unseen_wrn_28_2_combine}
    \end{subfigure}
    \hspace{1cm}
    \begin{subfigure}{.3\columnwidth}
      \centering
      \includegraphics[width=1\linewidth]{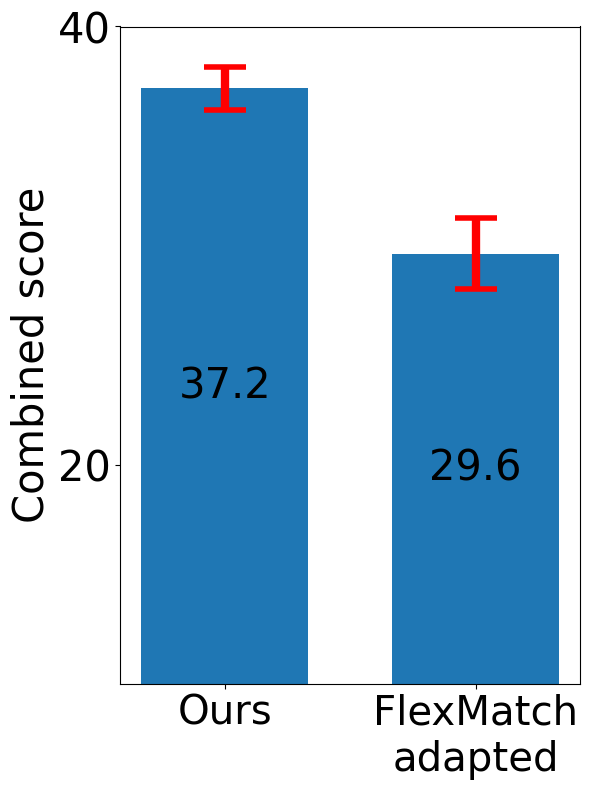}
      \vspace{-0.62cm}
      \caption{3 labels per class}
%      \vspace{0.2cm}
      \label{subfig:3_labels_per_class_35_unseen_wrn_28_2_combine}
    \end{subfigure}
    \caption{\textit{Combined score}, CIFAR-100, 35 unseen classes.}
    \label{fig:2_labels_per_class_35_unseen_wrn_28_2_combine}
      \vspace{-0.6cm}
\end{center}
\end{figure}

Fig.~\ref{fig:2_labels_per_class_35_unseen_wrn_28_2_combine} shows additional results with 35 unseen classes, further demonstrating the robustness of the method's beneficial contribution.

\section{Long-Tail Class Distribution}
\label{sec:long-tail}

\begin{figure}[htb]
\centering
    \begin{subfigure}{\columnwidth}
      \centering
      \includegraphics[width=.85\linewidth]{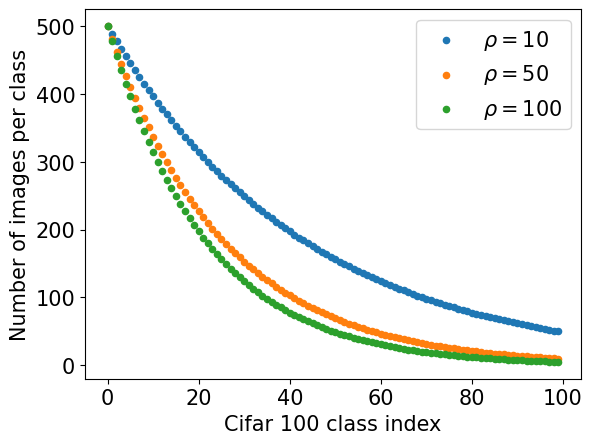}
    \end{subfigure}
    \caption{The distribution of classes in the long-tail CIFAR-100 is determined by $\rho$.} %\noam{adding 50, until we decide witch rho to use}}
    \label{fig:long_tail_graph}
\vspace{-0.6cm}
\end{figure}

\paragraph{Imbalanced CIFAR-100.} 
Following \citep{cao2019learning}, in order to obtain a long-tail class distribution of a dataset containing $C$ classes, the number of samples in each class $N_i$ is fixed at $N_i = N_{\text{max}} \cdot \rho^{-\frac{i}{C-1}}$. Here $\rho$ determines the imbalance ratio, and $N_0=N_{\text{max}}$ is the maximum number of samples per class. % $\rho = 1$ when labeled data is balanced over classes, and larger $\rho$ indicates more imbalanced class distribution. 

Fig.~\ref{fig:long_tail_graph} shows an example with $N_{\text{max}}=500$, which is the case for CIFAR-100, and 3 values of $\rho$. Accordingly, in Figs.~\ref{fig:long_tail_lt_ratio_10_total_sum_165}-\ref{fig:long_tail_lt_ratio_10_total_sum_260}, the term "modest long-tail distribution" corresponds to $\rho = 10$, while "sharp long-tail distribution" corresponds to $\rho = 50$ in Fig.~\ref{fig:long_tail_lt_ratio_50_total_sum_238} and to $\rho = 100$ in Fig.~\ref{fig:long_tail_lt_ratio_100_total_sum_247}.

In our experiments, we designated the unseen classes to be the first classes in the distribution, namely, classes $i\in[C_{unseen}]$ with a total of $\sum_{i=1}^{C_{unseen}} N_i$ examples. To adapt SSL algorithms in this scenario, the rejection threshold is set to guarantee that in the model's predictions, the fraction of rejected examples matches the correct fraction as dictated by the long-tail class distribution.

% \begin{figure}[htb]
% \begin{center}
    
%     \begin{subfigure}{1\columnwidth}
%       \centering
%       \includegraphics[width=1\linewidth]{graphs/long_tail/lt_ratio_50/total_sum_238/f1_score_summarize_choose_unseen_count_by_test_1.png}
%       \vspace{-0.62cm}
%       \caption{}
% %      \vspace{0.2cm}
%       \label{subfig:f1_score_lt_ratio_50_total_sum_238}
%     \end{subfigure}
%     \caption{F1 score, long tail taio - 50, 40 unseen classes, 248 labeled data,\noam{only after 200 epochs +-, not final! our algorithm result will eventually be a biy lower}}
%     \label{fig:f1_score_lt_ratio_50_total_sum_238}
%       \vspace{-0.25cm}
% \end{center}
% \end{figure}
\end{document}

%% file: preamble.tex
\usepackage[dvipsnames]{xcolor}

%% file: CVPR2024.bbl
\begin{thebibliography}{31}
\providecommand{\natexlab}[1]{#1}
\providecommand{\url}[1]{\texttt{#1}}
\expandafter\ifx\csname urlstyle\endcsname\relax
  \providecommand{\doi}[1]{doi: #1}\else
  \providecommand{\doi}{doi: \begingroup \urlstyle{rm}\Url}\fi

\bibitem[Berthelot et~al.(2019)Berthelot, Carlini, Goodfellow, Papernot,
  Oliver, and Raffel]{berthelot2019mixmatch}
David Berthelot, Nicholas Carlini, Ian Goodfellow, Nicolas Papernot, Avital
  Oliver, and Colin~A Raffel.
\newblock Mixmatch: A holistic approach to semi-supervised learning.
\newblock \emph{Advances in neural information processing systems}, 32, 2019.

\bibitem[Cao et~al.(2019)Cao, Wei, Gaidon, Arechiga, and Ma]{cao2019learning}
Kaidi Cao, Colin Wei, Adrien Gaidon, Nikos Arechiga, and Tengyu Ma.
\newblock Learning imbalanced datasets with label-distribution-aware margin
  loss.
\newblock \emph{Advances in neural information processing systems}, 32, 2019.

\bibitem[Cao et~al.(2022)Cao, Brbic, and Leskovec]{cao2022open}
Kaidi Cao, Maria Brbic, and Jure Leskovec.
\newblock Open-world semi-supervised learning.
\newblock In \emph{International Conference on Learning Representations
  (ICLR)}, 2022.

\bibitem[Chapelle et~al.(2009)Chapelle, Scholkopf, and Zien]{chapelle2009semi}
Olivier Chapelle, Bernhard Scholkopf, and Alexander Zien.
\newblock Semi-supervised learning (chapelle, o. et al., eds.; 2006)[book
  reviews].
\newblock \emph{IEEE Transactions on Neural Networks}, 20\penalty0
  (3):\penalty0 542--542, 2009.

\bibitem[Chen et~al.(2020)Chen, Zhu, Li, and Gong]{DBLP:conf/aaai/ChenZLG20}
Yanbei Chen, Xiatian Zhu, Wei Li, and Shaogang Gong.
\newblock Semi-supervised learning under class distribution mismatch.
\newblock In \emph{The Thirty-Fourth {AAAI} Conference on Artificial
  Intelligence, {AAAI} 2020, The Thirty-Second Innovative Applications of
  Artificial Intelligence Conference, {IAAI} 2020, The Tenth {AAAI} Symposium
  on Educational Advances in Artificial Intelligence, {EAAI} 2020, New York,
  NY, USA, February 7-12, 2020}, pages 3569--3576. {AAAI} Press, 2020.

\bibitem[Chen et~al.(2022)Chen, Mancini, Zhu, and Akata]{chen2022semi}
Yanbei Chen, Massimiliano Mancini, Xiatian Zhu, and Zeynep Akata.
\newblock Semi-supervised and unsupervised deep visual learning: A survey.
\newblock \emph{IEEE transactions on pattern analysis and machine
  intelligence}, 2022.

\bibitem[Coates et~al.(2011)Coates, Ng, and Lee]{coates2011analysis}
Adam Coates, Andrew Ng, and Honglak Lee.
\newblock An analysis of single-layer networks in unsupervised feature
  learning.
\newblock In \emph{Proceedings of the fourteenth international conference on
  artificial intelligence and statistics}, pages 215--223. JMLR Workshop and
  Conference Proceedings, 2011.

\bibitem[Grandini et~al.(2020)Grandini, Bagli, and Visani]{grandini2020metrics}
Margherita Grandini, Enrico Bagli, and Giorgio Visani.
\newblock Metrics for multi-class classification: an overview.
\newblock \emph{arXiv preprint arXiv:2008.05756}, 2020.

\bibitem[Guo et~al.(2020)Guo, Zhang, Jiang, Li, and Zhou]{guo2020safe}
Lan-Zhe Guo, Zhen-Yu Zhang, Yuan Jiang, Yu-Feng Li, and Zhi-Hua Zhou.
\newblock Safe deep semi-supervised learning for unseen-class unlabeled data.
\newblock In \emph{International Conference on Machine Learning}, pages
  3897--3906. PMLR, 2020.

\bibitem[Guo et~al.(2022)Guo, Zhang, Wu, Shao, and Li]{guo2022robust}
Lan-Zhe Guo, Yi-Ge Zhang, Zhi-Fan Wu, Jie-Jing Shao, and Yu-Feng Li.
\newblock Robust semi-supervised learning when not all classes have labels.
\newblock \emph{Advances in Neural Information Processing Systems},
  35:\penalty0 3305--3317, 2022.

\bibitem[Han et~al.(2020)Han, Rebuffi, Ehrhardt, Vedaldi, and
  Zisserman]{han2020automatically}
K Han, SA Rebuffi, S Ehrhardt, A Vedaldi, and A Zisserman.
\newblock Automatically discovering and learning new visual categories with
  ranking statistics.
\newblock In \emph{Proceedings of the 8th Intennational Conference on Learning
  Representations, ICLR 2020}. Schloss Dagstuhl-Leibniz-Zentrum f{\"u}r
  Informatik, 2020.

\bibitem[Huang et~al.(2021)Huang, Fang, Chen, Chai, Wei, Wei, Lin, and
  Li]{huang2021trash}
Junkai Huang, Chaowei Fang, Weikai Chen, Zhenhua Chai, Xiaolin Wei, Pengxu Wei,
  Liang Lin, and Guanbin Li.
\newblock Trash to treasure: Harvesting ood data with cross-modal matching for
  open-set semi-supervised learning.
\newblock In \emph{Proceedings of the IEEE/CVF International Conference on
  Computer Vision}, pages 8310--8319, 2021.

\bibitem[Krizhevsky et~al.(2009)Krizhevsky, Hinton,
  et~al.]{krizhevsky2009learning}
Alex Krizhevsky, Geoffrey Hinton, et~al.
\newblock Learning multiple layers of features from tiny images.
\newblock 2009.

\bibitem[Li et~al.(2015)Li, Guo, and Schuurmans]{li2015semi}
Xin Li, Yuhong Guo, and Dale Schuurmans.
\newblock Semi-supervised zero-shot classification with label representation
  learning.
\newblock In \emph{Proceedings of the IEEE international conference on computer
  vision}, pages 4211--4219, 2015.

\bibitem[Nguyen et~al.(2015)Nguyen, Yosinski, and Clune]{nguyen2015deep}
Anh Nguyen, Jason Yosinski, and Jeff Clune.
\newblock Deep neural networks are easily fooled: High confidence predictions
  for unrecognizable images.
\newblock In \emph{Proceedings of the IEEE conference on computer vision and
  pattern recognition}, pages 427--436, 2015.

\bibitem[Park et~al.(2022)Park, Yun, Jeong, and Shin]{park2022opencos}
Jongjin Park, Sukmin Yun, Jongheon Jeong, and Jinwoo Shin.
\newblock Opencos: Contrastive semi-supervised learning for handling open-set
  unlabeled data.
\newblock In \emph{European Conference on Computer Vision}, pages 134--149.
  Springer, 2022.

\bibitem[Pourpanah et~al.(2022)Pourpanah, Abdar, Luo, Zhou, Wang, Lim, Wang,
  and Wu]{pourpanah2022review}
Farhad Pourpanah, Moloud Abdar, Yuxuan Luo, Xinlei Zhou, Ran Wang, Chee~Peng
  Lim, Xi-Zhao Wang, and QM~Jonathan Wu.
\newblock A review of generalized zero-shot learning methods.
\newblock \emph{IEEE transactions on pattern analysis and machine
  intelligence}, 2022.

\bibitem[Rizve et~al.(2022)Rizve, Kardan, Khan, Shahbaz~Khan, and
  Shah]{rizve2022openldn}
Mamshad~Nayeem Rizve, Navid Kardan, Salman Khan, Fahad Shahbaz~Khan, and
  Mubarak Shah.
\newblock Openldn: Learning to discover novel classes for open-world
  semi-supervised learning.
\newblock In \emph{European Conference on Computer Vision}, pages 382--401.
  Springer, 2022.

\bibitem[Saito et~al.(2021)Saito, Kim, and
  Saenko]{DBLP:journals/corr/abs-2105-14148}
Kuniaki Saito, Donghyun Kim, and Kate Saenko.
\newblock Openmatch: Open-set consistency regularization for semi-supervised
  learning with outliers.
\newblock \emph{CoRR}, abs/2105.14148, 2021.

\bibitem[Sohn et~al.(2020)Sohn, Berthelot, Carlini, Zhang, Zhang, Raffel,
  Cubuk, Kurakin, and Li]{sohn2020fixmatch}
Kihyuk Sohn, David Berthelot, Nicholas Carlini, Zizhao Zhang, Han Zhang,
  Colin~A Raffel, Ekin~Dogus Cubuk, Alexey Kurakin, and Chun-Liang Li.
\newblock Fixmatch: Simplifying semi-supervised learning with consistency and
  confidence.
\newblock \emph{Advances in neural information processing systems},
  33:\penalty0 596--608, 2020.

\bibitem[Sun and Li(2023)]{DBLP:journals/tmlr/SunL23}
Yiyou Sun and Yixuan Li.
\newblock Opencon: Open-world contrastive learning.
\newblock \emph{Trans. Mach. Learn. Res.}, 2023, 2023.

\bibitem[Tax and Duin(2008)]{tax2008growing}
David~MJ Tax and Robert~PW Duin.
\newblock Growing a multi-class classifier with a reject option.
\newblock \emph{Pattern Recognition Letters}, 29\penalty0 (10):\penalty0
  1565--1570, 2008.

\bibitem[Wang et~al.(2022)Wang, Chen, Fan, Sun, Tao, Hou, Wang, Yang, Zhou,
  Guo, Qi, Wu, Li, Nakamura, Ye, Savvides, Raj, Shinozaki, Schiele, Wang, Xie,
  and Zhang]{usb2022}
Yidong Wang, Hao Chen, Yue Fan, Wang Sun, Ran Tao, Wenxin Hou, Renjie Wang,
  Linyi Yang, Zhi Zhou, Lan-Zhe Guo, Heli Qi, Zhen Wu, Yu-Feng Li, Satoshi
  Nakamura, Wei Ye, Marios Savvides, Bhiksha Raj, Takahiro Shinozaki, Bernt
  Schiele, Jindong Wang, Xing Xie, and Yue Zhang.
\newblock Usb: A unified semi-supervised learning benchmark for classification.
\newblock In \emph{Thirty-sixth Conference on Neural Information Processing
  Systems Datasets and Benchmarks Track}, 2022.

\bibitem[Wang et~al.(2023)Wang, Chen, Heng, Hou, Fan, Wu, Wang, Savvides,
  Shinozaki, Raj, Schiele, and Xie]{DBLP:conf/iclr/Wang0HHFW0SSRS023}
Yidong Wang, Hao Chen, Qiang Heng, Wenxin Hou, Yue Fan, Zhen Wu, Jindong Wang,
  Marios Savvides, Takahiro Shinozaki, Bhiksha Raj, Bernt Schiele, and Xing
  Xie.
\newblock Freematch: Self-adaptive thresholding for semi-supervised learning.
\newblock In \emph{The Eleventh International Conference on Learning
  Representations, {ICLR} 2023, Kigali, Rwanda, May 1-5, 2023}. OpenReview.net,
  2023.

\bibitem[Wang et~al.(2020)Wang, Salehi, Gritsenko, Chowdhury, Ioannidis, and
  Dy]{wang2020open}
Zifeng Wang, Batool Salehi, Andrey Gritsenko, Kaushik Chowdhury, Stratis
  Ioannidis, and Jennifer Dy.
\newblock Open-world class discovery with kernel networks.
\newblock In \emph{2020 IEEE International Conference on Data Mining (ICDM)},
  pages 631--640. IEEE, 2020.

\bibitem[Xie et~al.(2016)Xie, Girshick, and Farhadi]{xie2016unsupervised}
Junyuan Xie, Ross Girshick, and Ali Farhadi.
\newblock Unsupervised deep embedding for clustering analysis.
\newblock In \emph{International conference on machine learning}, pages
  478--487. PMLR, 2016.

\bibitem[Xu et~al.(2021)Xu, Zeng, Lian, and Ding]{xu2021semi}
Bingrong Xu, Zhigang Zeng, Cheng Lian, and Zhengming Ding.
\newblock Semi-supervised low-rank semantics grouping for zero-shot learning.
\newblock \emph{IEEE Transactions on Image Processing}, 30:\penalty0
  2207--2219, 2021.

\bibitem[Yu et~al.(2020)Yu, Ikami, Irie, and Aizawa]{yu2020multi}
Qing Yu, Daiki Ikami, Go Irie, and Kiyoharu Aizawa.
\newblock Multi-task curriculum framework for open-set semi-supervised
  learning.
\newblock In \emph{Computer Vision--ECCV 2020: 16th European Conference,
  Glasgow, UK, August 23--28, 2020, Proceedings, Part XII 16}, pages 438--454.
  Springer, 2020.

\bibitem[Zagoruyko and Komodakis(2016)]{zagoruyko2016wide}
Sergey Zagoruyko and Nikos Komodakis.
\newblock Wide residual networks.
\newblock \emph{arXiv preprint arXiv:1605.07146}, 2016.

\bibitem[Zhang et~al.(2021)Zhang, Wang, Hou, Wu, Wang, Okumura, and
  Shinozaki]{DBLP:conf/nips/ZhangWHWWOS21}
Bowen Zhang, Yidong Wang, Wenxin Hou, Hao Wu, Jindong Wang, Manabu Okumura, and
  Takahiro Shinozaki.
\newblock Flexmatch: Boosting semi-supervised learning with curriculum pseudo
  labeling.
\newblock In \emph{Advances in Neural Information Processing Systems 34: Annual
  Conference on Neural Information Processing Systems 2021, NeurIPS 2021,
  December 6-14, 2021, virtual}, pages 18408--18419, 2021.

\bibitem[Zhong et~al.(2021)Zhong, Zhu, Luo, Li, Yang, and
  Sebe]{zhong2021openmix}
Zhun Zhong, Linchao Zhu, Zhiming Luo, Shaozi Li, Yi Yang, and Nicu Sebe.
\newblock Openmix: Reviving known knowledge for discovering novel visual
  categories in an open world.
\newblock In \emph{Proceedings of the IEEE/CVF Conference on Computer Vision
  and Pattern Recognition}, pages 9462--9470, 2021.

\end{thebibliography}
